%% file: wileyNJDv5_AMA.tex
\documentclass[AMA,Times1COL]{WileyNJDv5} 

\received{Date Month Year}
\revised{Date Month Year}
\accepted{Date Month Year}
\journal{Journal}
\volume{00}
\copyyear{2026}
\startpage{1}

\usepackage[utf8]{inputenc}
\usepackage[T1]{fontenc}

\usepackage{amsmath}

\usepackage[table,svgnames]{xcolor}
\usepackage{booktabs} 
\usepackage{threeparttable}
\usepackage{multirow} 
\usepackage{tabularx} 
\usepackage{array}    
\usepackage{float}    
\usepackage{adjustbox}

\usepackage{soul}
\sethlcolor{yellow}   

\usepackage{graphicx} 
\usepackage{tikz}     
\usetikzlibrary{tikzmark, calc, arrows.meta, shapes.callouts} 
\usetikzlibrary{trees, positioning, shapes} 

\usepackage{tcolorbox}
\tcbuselibrary{breakable} 
\usepackage{enumitem}

\usepackage{hyperref}

\newlist{criteria}{enumerate}{1}
\setlist[criteria]{label=\arabic*., leftmargin=*, itemsep=2pt}

\makeatletter
\def\Authorfont{\rmfamily\fontsize{12}{18}\selectfont\bfseries\raggedright}%
\makeatother

\begin{document}

\title{Medical Reasoning with Large Language Models: A Survey and MR-Bench}

\author[1]{Xiaohan Ren}

\author[2]{Chenxiao Fan}

\author[2]{Wenyin Ma}

\author[3]{Hongliang He}

\author[4]{Chongming Gao*}

\author[5]{Xiaoyan Zhao*}

\author[2]{Fuli Feng*}

\authormark{REN \textsc{et al.}}
\titlemark{Medical Reasoning with Large Language Models: A Survey and MR-Bench}

\address[1]{\orgdiv{School of Cyber Science and Technology}, \orgname{University of Science and Technology of China}, \orgaddress{\state{Anhui}, \country{China}}}

\address[2]{\orgdiv{School of Artificial Intelligence and Data Science}, \orgname{University of Science and Technology of China}, \orgaddress{\state{Anhui}, \country{China}}}

\address[3]{\orgdiv{Department of Infectious Disease}, \orgname{The First Affiliated Hospital of USTC}, \orgaddress{\state{Anhui}, \country{China}}}

\address[4]{\orgdiv{School of Information Science and Technology}, \orgname{University of Science and Technology of China}, \orgaddress{\state{Anhui}, \country{China}}}

\address[5]{\orgdiv{School of Computing}, \orgname{National University of Singapore}, \orgaddress{\country{Singapore}}}

\corres{Corresponding authors: Chongming Gao, Xiaoyan Zhao, and Fuli Feng. \email{chongming.gao@gmail.com}; \email{xzhao@se.cuhk.edu.hk}; \email{fulifeng93@gmail.com}}



\abstract[Abstract]{Large language models (LLMs) have achieved strong performance on medical exam-style tasks, motivating growing interest in their deployment in real-world clinical settings. However, clinical decision-making is inherently safety-critical, context-dependent, and conducted under evolving evidence. In such situations, reliable LLM performance depends not on factual recall alone, but on robust medical reasoning.
In this work, we present a comprehensive review of medical reasoning with LLMs. Grounded in cognitive theories of clinical reasoning, we conceptualize medical reasoning as an iterative process of abduction, deduction, and induction, and organize existing methods into seven major technical routes spanning training-based and training-free approaches. We further conduct a unified cross-benchmark evaluation of representative medical reasoning models under a consistent experimental setting, enabling a more systematic and comparable assessment of the empirical impact of existing methods.
To better assess clinically grounded reasoning, we introduce MR-Bench, a benchmark derived from real-world hospital data. Evaluations on MR-Bench expose a pronounced gap between exam-level performance and accuracy on authentic clinical decision tasks. Overall, this survey provides a unified view of existing medical reasoning methods, benchmarks, and evaluation practices, and highlights key gaps between current model performance and the requirements of real-world clinical reasoning.\newline\textbf{Code:} \url{https://github.com/RXH04-USTC/Medical-Reasoning-Survey}}
\keywords{clinical decision making; cross-benchmark evaluation; large language models; medical reasoning}

\jnlcitation{\cname{%
\author{Ren X.},
\author{Fan C.},
\author{Ma W.},
\author{Gao C.},
\author{Zhao X.}, and
\author{Feng F.}}
\ctitle{Medical Reasoning with Large Language Models: A Systematic Review and Evaluation.} \cjournal{\it iNew Medicine} \cvol{2026;00(00):1--XX}.}

\maketitle



\input{sections/intro}

\input{sections/sec2}

\input{sections/sec3_benchmarks}

\section{MR-Bench: A Real-World Clinical Benchmark}\label{section4}

\subsection{Limitation of Existing Benchmarks}
Despite the proliferation of evaluation datasets, current medical benchmarks still exhibit certain limitations. First, widely adopted QA datasets such as MedQA \cite{jin2021disease}, PubMedQA \cite{jin2019pubmedqa}, and MedMCQA \cite{pal2022medmcqa} are primarily derived from medical licensing examinations or literature. While valuable, these sources may diverge from the complexities of authentic clinical environments. Conversely, benchmarks rooted in clinical data face different challenges; for instance, the risk prediction tasks in EHR-Bench \cite{liao2025ehr} rely on token probability-based evaluation, which may inherently disadvantage models that have not undergone specific training for such outputs \cite{zhao2021calibrate}. Similarly, while EHRNoteQA \cite{kweon2024ehrnoteqa} attempts to simulate clinical settings by extracting QA tasks from EHRs, its reported baselines demonstrate that advanced models like GPT-4 already achieve scores exceeding 90\%, suggesting that the benchmark may lack the necessary challenge to distinguish frontier capabilities. Furthermore, existing benchmarks are rarely designed specifically to evaluate medical reasoning. Although datasets like MedXpertQA \cite{zuo2025medxpertqa} and MedAgentsBench \cite{anonymous2025medagentsarena} increase difficulty by filtering samples based on LLM pass rates, they fail to explicitly establish a connection between these hard samples and specific medical reasoning sub-capabilities.

To provide a more concrete understanding of these limitations, we conduct a granular analysis that reveals two representative and recurring issues in current medical benchmarks.

\noindent \textbf{Issue 1: Data Quality and Information Completeness.}
A fundamental limitation of widely used medical QA benchmarks lies in data quality and information completeness. Taking MedQA \cite{jin2021disease} as a representative example, this dataset—introduced in 2021 and formatted after the USMLE—has become a cornerstone benchmark for evaluating medical LLMs. However, recent investigations, such as those by the Med-Gemini team \cite{saab2024capabilities}, have identified critical flaws, including missing key information, label errors, and ambiguous questions.

Our own analysis corroborates these findings and further reveals that \emph{missing contextual information} is both significant and prevalent. In particular, many questions explicitly reference external visual evidence (e.g., ``as shown in the image'') without providing the corresponding images. Such omissions undermine the validity of the evaluation, as models are implicitly penalized for failing to reason over information that is not available to them. These issues point to insufficient data auditing and raise concerns about whether performance on such benchmarks genuinely reflects medical reasoning ability. Representative cases are provided in Appendix~\ref{Appendix C.1}.

\noindent \textbf{Issue 2: Instability and Subjectivity in LLM-Based Evaluation.}
Another critical limitation arises from the growing reliance on LLM-based evaluators for open-ended medical reasoning benchmarks. HealthBench \cite{arora2025healthbench} exemplifies this evaluation paradigm by adopting a multi-criteria assessment framework in which human experts define scoring rubrics and a designated judge model—officially GPT-4.1—determines whether a model's response satisfies these criteria.

However, our experiments demonstrate that such evaluations are highly sensitive to the choice of the judge model. Using Qwen2.5-7B-Instruct as the test model, we evaluated a subset of 500 HealthBench samples under identical generation outputs but with different judges. When evaluated by the official GPT-4.1 judge, the model achieved a score of 0.247 (SD=0.016), whereas substituting the judge with GPT-4.1-mini yielded a substantially higher score of 0.420 (SD=0.015). Since the generated responses were held constant across runs, this discrepancy can only be attributed to evaluator variability.

These results expose a fundamental instability in current LLM-as-a-judge evaluation pipelines and call into question their reliability as objective measures of medical reasoning. Without stronger guarantees of evaluator consistency and calibration, such benchmarks risk conflating model capability with evaluator bias. Detailed analyses and examples are provided in Appendix~\ref{Appendix C.2}.

\subsection{Benchmark Construction}


The limitations of existing medical benchmarks reveal three key gaps: insufficient grounding in authentic clinical decision contexts, limited evaluation of clinically meaningful reasoning beyond answer correctness, and instability in open-ended evaluation protocols. Exam-style QA benchmarks abstract away real-world constraints, while clinical benchmarks relying on LLM-based judges are sensitive to evaluator choice. These issues motivate the construction of a benchmark that targets real-world medical reasoning with stable and reproducible evaluation.

We therefore introduce \textbf{MR-Bench}, a medical reasoning benchmark derived entirely from authentic electronic health records. MR-Bench formulates evaluation as clinical decision-making under incomplete information and safety constraints, rather than generic question answering. To this end, we design two representative decision tasks—\textit{Medication Imputation} and \textit{Procedure Selection}—which correspond to high-impact actions routinely performed in clinical workflows.

Both tasks are constructed to require the joint application of \emph{medical deduction} and \emph{medical induction} (Section~\ref{section1}). Models must employ medical deduction to assess candidate decisions against clinical guidelines, contraindications, and drug--drug interactions, while using medical induction to synthesize heterogeneous patient evidence, including diagnoses, medical history, and clinical notes, to determine appropriate interventions.

To mitigate the evaluation instability observed in open-ended benchmarks, we adopt a multiple-choice question (MCQ) formulation. This choice provides an objective and reproducible evaluation signal while preserving reasoning complexity through carefully constructed, clinically grounded distractors.



\subsubsection{Data Source and Temporal Generalization}

To evaluate models under evolving clinical practices rather than a single static snapshot, MR-Bench is constructed to explicitly test \textit{temporal generalization}. Specifically, we aim to assess whether models can maintain reliable medical reasoning across changes in clinical guidelines, drug approval standards, and documentation practices over time.

To this end, we extract data from MIMIC-IV, a large-scale, de-identified clinical database containing laboratory results, medication prescriptions, and clinical notes from intensive care patients. We preserve the original temporal stratification of the dataset and sample records from four distinct time periods. From each period, we randomly select 250 complete hospital admissions, yielding a total of 1,000 high-quality clinical cases. This temporal sampling strategy ensures that evaluation results reflect robust generalization across decades rather than overfitting to the medical patterns of a specific era.

\subsubsection{Clinical Representation Engineering}

To enable reliable medical reasoning rather than surface-level pattern matching, MR-Bench requires clinical representations that are both information-sufficient and free from spurious cues or outcome leakage. We therefore design a representation engineering pipeline that transforms heterogeneous and noisy raw medical records into concise, decision-relevant inputs suitable for LLM-based reasoning, drawing on the principles established in LAMO \cite{LAMO} and FLAME \cite{FLAME}.

For structured clinical information, we perform semantic refinement and filtering to balance coverage and interpretability. ICD diagnosis codes, procedure codes, and medication lists are retrieved via the $hadm\_id$, and long-tail items are excluded to reduce noise from infrequent cases. In addition, the original ICD \textit{long titles} are often excessively verbose and cognitively burdensome for language models. To mitigate this issue, we employ GPT-4o to rewrite these entries into concise and readable titles, preserving clinical meaning while substantially reducing linguistic complexity (e.g., rewriting lengthy ICD descriptions into compact disease names).

For unstructured clinical notes, our goal is to provide models with realistic admission-time information while preventing information leakage from future outcomes. Prior work has shown that directly using discharge summaries can inadvertently expose final diagnoses and treatments. To address this, we extract four core clinical elements from progress notes using GPT-4o: \textit{History of Present Illness} (HPI), \textit{Past Medical History} (PMH), \textit{Allergic History}, and \textit{Medication on Admission}. Together, these elements form the natural language representation of each patient, closely mirroring the information available to clinicians during the early stages of clinical decision-making.

\subsubsection{Task Design and Distractor Construction}

To evaluate decision-oriented medical reasoning under realistic clinical constraints, we design two 8-choice multiple-answer tasks that require models to select appropriate clinical actions based on incomplete patient information and safety considerations.

\begin{enumerate}
    \item \textbf{Medication Imputation.}
    This task assesses whether a model can infer an appropriate medication decision given partial prescription information. Models are provided with diagnosis titles, procedure titles, and four natural language clinical elements, while the prescription list is partially masked. To prevent shortcut reasoning and explicitly evaluate pharmacological safety awareness, distractor medications are preferentially sampled from the TWOSIDES dataset~\cite{ddi} to include options with known drug--drug interaction risks or contraindications with the correct answer; remaining options are sampled from the general medication pool.

    \item \textbf{Procedure Selection.}
    This task evaluates a model’s ability to translate clinical evidence into appropriate interventional decisions. Given the patient’s diagnoses and natural language representations, the model must select the correct set of procedures for the admission. Distractors are sampled from procedures associated with other patients, ensuring plausibility while remaining clinically inappropriate, thereby testing the model’s ability to form a coherent ``clinical manifestation-to-intervention'' reasoning chain.
\end{enumerate}

\begin{table}[!htbp]
    \centering
    \caption{Model performance on MR-Bench. Background colors denote base LLM families, with each medical LLM inheriting the background color of its corresponding base model. \textbf{Bold} indicates the best overall score (excluding advanced fundation models); \protect\underline{underlining} marks the best medical LLM within each base model group. R.I. denotes Relative Improvement compared to the corresponding base model.}
    \label{tab:drug_procedure_performance}
    \renewcommand{\arraystretch}{1.35} 
    
    \fontsize{8pt}{9pt}\selectfont
    
    \newcolumntype{Y}{>{\centering\arraybackslash}X}
    
    \begin{tabularx}{\textwidth}{ l Y Y Y Y Y Y }
        \toprule
        \multirow{2}{*}{\textbf{Model}} & \multicolumn{2}{c}{\textbf{Medication}} & \multicolumn{2}{c}{\textbf{Procedure}} & \multicolumn{2}{c}{\textbf{Average}} \\
        \cmidrule(lr){2-3} \cmidrule(lr){4-5} \cmidrule(lr){6-7}
         & \textbf{Acc.} & \textbf{R.I. (\%)} & \textbf{Acc.} & \textbf{R.I. (\%)} & \textbf{Acc.} & \textbf{R.I. (\%)} \\
        \midrule
        \multicolumn{7}{c}{\textit{Open-source General LLMs (Base Models)}} \\
        \midrule
        \rowcolor{LightBlue!30} gemma3-4B-it & 0.237 & - & 0.504 & - & 0.371 & - \\
        Qwen3-4B-Thinking-2507 & 0.272 & - & 0.588 & - & 0.430 & - \\
        \rowcolor{LightYellow} Qwen-2.5-7B-Instruct & 0.213 & - & 0.595 & - & 0.404 & - \\
        \rowcolor{Lavender} Meta-Llama-3-8B-Instruct & 0.166 & - & 0.431 & - & 0.299 & - \\
        \rowcolor{LightGray!30} Llama-3.1-8B-Instruct & 0.153 & - & 0.416 & - & 0.285 & - \\
        Qwen3-8B & 0.269 & - & \textbf{0.602} & - & \textbf{0.436} & - \\
        \midrule
        \multicolumn{7}{c}{\textit{Open-source Medical LLMs (vs. Base General LLM)}} \\
        \midrule
        \rowcolor{LightBlue!30} Medgemma-4B & \protect\underline{0.214} & \textcolor{red}{-9.70} & \protect\underline{0.324} & \textcolor{red}{-35.71} & \protect\underline{0.269} & \textcolor{red}{-27.36} \\
        
        \rowcolor{LightYellow} ClinicalGPT-R1-Qwen-7B & \protect\underline{0.225} & \textcolor{ForestGreen}{+5.63} & 0.417 & \textcolor{red}{-29.92} & 0.321 & \textcolor{red}{-20.54} \\
        
        \rowcolor{LightYellow} HuatuoGPT-o1-7B & 0.220 & \textcolor{ForestGreen}{+3.29} & \protect\underline{0.599} & \textcolor{ForestGreen}{+0.67} & \protect\underline{0.410} & \protect\underline{\textcolor{ForestGreen}{+1.49}} \\
        
        \rowcolor{LightYellow} m1-7B-23K & 0.206 & \textcolor{red}{-3.29} & 0.575 & \textcolor{red}{-3.36} & 0.391 & \textcolor{red}{-3.22} \\
        
        \rowcolor{Lavender} Med42-v2-8B & \protect\underline{0.249} & \protect\underline{\textcolor{ForestGreen}{+50.00}} & \protect\underline{0.585} & \protect\underline{\textcolor{ForestGreen}{+35.73}} & \protect\underline{0.417} & \protect\underline{\textcolor{ForestGreen}{+39.67}} \\
        \rowcolor{Lavender} Meerkat-8B & 0.219 & \textcolor{ForestGreen}{+31.93} & 0.539 & \textcolor{ForestGreen}{+25.06} & 0.379 & \textcolor{ForestGreen}{+26.96} \\
        
        \rowcolor{LightGray!30} Adathink-Med-Llama-8B & 0.226 & \textcolor{ForestGreen}{+47.71} & 0.574 & \textcolor{ForestGreen}{+37.98} & 0.400 & \textcolor{ForestGreen}{+40.60} \\
        \rowcolor{LightGray!30} AlphaMed-8B & 0.210 & \textcolor{ForestGreen}{+37.25} & 0.532 & \textcolor{ForestGreen}{+27.88} & 0.371 & \textcolor{ForestGreen}{+30.40} \\
        \rowcolor{LightGray!30} BioMed-R1-8B & 0.242 & \textcolor{ForestGreen}{+58.17} & 0.552 & \textcolor{ForestGreen}{+32.69} & 0.397 & \textcolor{ForestGreen}{+39.54} \\
        \rowcolor{LightGray!30} HuatuoGPT-o1-8B & 0.225 & \textcolor{ForestGreen}{+47.06} & 0.521 & \textcolor{ForestGreen}{+25.24} & 0.373 & \textcolor{ForestGreen}{+31.11} \\
        \rowcolor{LightGray!30} Llama-3.1-8B-UltraMedical & 0.231 & \textcolor{ForestGreen}{+50.98} & \protect\underline{0.586} & \protect\underline{\textcolor{ForestGreen}{+40.87}} & 0.409 & \textcolor{ForestGreen}{+43.59} \\
        \rowcolor{LightGray!30} MedReason-8B & 0.220 & \textcolor{ForestGreen}{+43.79} & 0.482 & \textcolor{ForestGreen}{+15.87} & 0.351 & \textcolor{ForestGreen}{+23.37} \\
        \rowcolor{LightGray!30} MedS$^3$-8B & \protect\underline{\textbf{0.287}} & \protect\underline{\textcolor{ForestGreen}{+87.58}} & 0.578 & \textcolor{ForestGreen}{+38.94} & \protect\underline{0.433} & \protect\underline{\textcolor{ForestGreen}{+52.02}} \\
        \midrule
        
        \multicolumn{7}{c}{\textit{Advanced Foundation Models}} \\
        \midrule
        Deepseek-V3.2 & 0.484 & - & 0.694 & - & 0.589 & - \\
        Gemini-3.0-pro & 0.550 & - & 0.676 & - & 0.613 & - \\
        GPT-4o & 0.363 & - & 0.649 & - & 0.506 & - \\
        GPT-5 & 0.544 & - & 0.738 & - & 0.641 & - \\
        Qwen3-max & 0.446 & - & 0.682 & - & 0.564 & - \\
        \bottomrule
    \end{tabularx}
\begin{tablenotes}
\item \textit{Abbreviations}: R.I.: Relative Improvement (compared to the corresponding base general LLM); \textit{Acc}.: Accuracy.
\end{tablenotes} 
\end{table}

\subsection{Evaluation and Analysis}

We evaluate MR-Bench on a diverse set of open-source medical and general-purpose models (Table~\ref{tab:model_performance}), together with advanced foundation models, including Deepseek-V3.2 (thinking), Gemini-3.0-pro (thinking), GPT-4o, GPT-5 (thinking), and Qwen3-max. All models are evaluated with a temperature of 0.5 for stability, and results are reported in Table~\ref{tab:drug_procedure_performance}.

Among open-source models with fewer than 10B parameters, the general-purpose Qwen3-8B achieves the highest average accuracy (0.436), while MedS$^{3}$-8B performs best on the medication imputation subtask. Notably, we observe a systematic divergence from standard medical benchmark results: several medical fine-tuned models that show strong gains on exam-style datasets such as MedQA (e.g., Medgemma-4B and m1-7B-23K) exhibit performance regression on MR-Bench relative to their base models. This pattern suggests that performance on MR-Bench is jointly determined by base model capacity and the alignment between tuning data and authentic clinical reasoning demands.

This effect is further illustrated by comparing identical tuning strategies across different base models. Because Qwen-2.5 substantially outperforms Llama-3.1 on MR-Bench (0.404 vs. 0.285), HuatuoGPT-o1 yields a +31.11\% gain when built on Llama, but only a marginal +1.49\% improvement when built on Qwen, despite achieving a higher absolute score. These results indicate that current medical tuning datasets—largely derived from exams or synthetic data—may inadequately reflect the complexity of real-world clinical records, limiting their ability to consistently enhance base model reasoning.

In contrast, advanced foundation models substantially outperform smaller specialized models, confirming that scale and extensive pretraining remain dominant factors for complex medical reasoning. GPT-5 achieves the highest overall accuracy (0.641) and leads in procedure selection, while Gemini-3.0-pro performs best on medication imputation. Nevertheless, even these frontier models only reach approximately a 60\% accuracy level on MR-Bench, highlighting persistent challenges in temporal reasoning and evidence synthesis and underscoring the gap between current LLM capabilities and deployment-ready clinical reliability.

\section{Discussion and Future Directions}\label{section5}

Despite remarkable advancements, achieving truly clinically applicable medical reasoning with LLMs faces formidable challenges. In this section, we discuss existing limitations and outline future directions across four critical dimensions: restructuring evaluation standards, ensuring reliable reasoning processes, enabling dynamic active interaction, and addressing ethical and safety concerns.

\subsection{Toward Clinically Grounded Evaluation}

As medical reasoning LLMs advance, static benchmarks derived from medical licensing examinations (e.g., MedQA~\cite{jin2021disease}) are increasingly inadequate as proxies for real-world clinical utility. While such datasets remain useful for probing foundational knowledge, they fail to capture the dynamic, decision-driven nature of clinical practice.

Recent benchmarks have begun to address complementary aspects of this gap. HealthBench~\cite{arora2025healthbench} emphasizes medical safety through multi-dimensional harm-oriented evaluation, MedR-Bench~\cite{qiu2025quantifying} and EHR-Bench~\cite{liao2025ehr} operationalize longitudinal clinical workflows via predictive tasks on EHR sequences, and PHA-Bench~\cite{heydari2025anatomy} focuses on patient-centric, agentic health management. In this landscape, MR-Bench anchors evaluation in authentic EHR-derived scenarios and assesses whether models can transform complex clinical context into evidence-based medical decisions.

Despite these advances, current benchmarks largely evaluate outcomes rather than the underlying reasoning processes. A key future direction is to develop evaluation frameworks that explicitly disentangle and measure the core cognitive components of medical reasoning—medical abduction, deduction, and induction (Section~\ref{section1}). Achieving this goal will likely require hybrid evaluation paradigms that combine expert review, scalable LLM-based judging, and rule-based metrics (Section~\ref{section3}), balancing reliability, coverage, and efficiency.

\subsection{Active Interaction and Tool-Augmented Reasoning}

Clinical reasoning is inherently dynamic and longitudinal, involving iterative hypothesis updating as new evidence becomes available. In real-world settings, critical information is rarely complete upfront; clinicians reduce uncertainty by actively acquiring targeted evidence, such as asking discriminative follow-up questions or ordering confirmatory tests. To operate in authentic clinical environments, medical LLMs must therefore move beyond static analysis and learn to reason over evolving EHR trajectories, identifying high-impact information gaps and prioritizing what to query or verify next \cite{GAO2021100}.

Recent benchmarks have begun to probe this capability. For instance, MedR-Bench~\cite{qiu2025quantifying} evaluates multi-stage clinical workflows that require intermediate evidence-seeking decisions. However, current results \cite{qiu2025quantifying, tang2024medagents} indicate that LLMs remain weak at proactive information acquisition: they frequently miss key evidence, ask low-yield questions, and prematurely commit to diagnoses or plans. Improving such clinical ``information foraging''—under constraints of safety, cost, and invasiveness—should thus be treated as a core research objective.

In parallel, robust tool use is essential for reliable deployment. Retrieval tools (e.g., clinical guidelines or curated knowledge bases) can ground reasoning in up-to-date evidence, while computation tools (e.g., dosage calculators or risk scores) provide exact and auditable outputs. More broadly, tool invocation should be integrated into the reasoning loop: models must learn when to retrieve, when to compute, and when to request additional patient data, and to transparently attribute conclusions to the supporting external evidence.

\subsection{Toward Trustworthy Medical Reasoning}

For clinical deployment, medical reasoning LLMs must be not only accurate but trustworthy. Plausible chains of thought are insufficient: in high-stakes settings, errors in intermediate reasoning can justify unsafe actions, obscure uncertainty, or induce unwarranted confidence. Trustworthy medical reasoning therefore requires evidence-grounded decision making, calibrated uncertainty, and robustness against hallucinated rationales, rather than optimization for benchmark accuracy alone.

A key shift is to move from eliciting longer reasoning traces toward constraining and verifying them. Although paradigms such as o1~\cite{huang2025o1} and DeepSeek-R1~\cite{guo2025deepseekr1} show that increased inference-time computation can improve reasoning structure, longer reasoning is not inherently more reliable. Trustworthy systems should ground critical intermediate claims in authoritative external evidence (e.g., clinical guidelines or curated knowledge bases), explicitly distinguish supported steps from uncertain ones, and adopt fail-safe behaviors such as abstention under insufficient evidence, appropriate escalation, and harm-minimizing decision preferences.

Trustworthiness is further limited by the training signal. Real-world EHRs record outcomes rather than the clinical rationale, making direct supervision of abduction, deduction, and induction difficult. While recent work synthesizes reasoning traces with LLMs~\cite{kim2025small, wu2025medreason, chen2412huatuogpt}, verification remains the core challenge: without rigorous clinical auditing, models may learn fluent but medically invalid logic \cite{thirunavukkarasu2023large}. Progress thus depends on stronger process supervision, including expert review, guideline-based consistency checks, and trace-level filtering of unsupported assertions.

Finally, trustworthiness encompasses safety, bias, and ethics. Confident hallucination \cite{lee_bubeck_petro_2023} remains a primary safety risk and must be addressed through uncertainty-aware modeling and harm-oriented evaluation that tests failure modes such as missed contraindications or unsafe recommendations. In parallel, biases arising from skewed training data and health-system specificity necessitate systematic subgroup evaluation and transparent reporting \cite{omiye2023large}. Critically, deployment frameworks should position medical LLMs as decision-support tools rather than autonomous agents, preserving clinician accountability, evidence traceability, and robust privacy and governance guarantees.



\section{Conclusions}\label{section6}

This survey systematically reviews the landscape of medical reasoning LLMs, categorizing methodologies into training-based and training-free paradigms. Our evaluation demonstrates that while domain adaptation effectively enhances performance on standard examinations, these gains fail to translate to the authentic clinical challenges presented in our \textbf{MR-Bench}, where even frontier foundation models exhibit significant reasoning gaps. Consequently, future research must move beyond static benchmarks toward developing verifiable, active, and clinically robust reasoning systems.

\bmsection*{Author contributions}

\textbf{Xiaohan Ren}: Conceptualization, Methodology, Software, Validation, Formal analysis, Investigation, Data Curation, Writing - original draft, Visualization.\\
\textbf{Chenxiao Fan}: Conceptualization, Methodology, Software, Validation, Formal analysis, Investigation, Data Curation, Writing - original draft.\\
\textbf{Wenyin Ma}: Conceptualization, Methodology, Software, Validation, Formal analysis, Investigation, Data Curation, Writing - original draft.\\
\textbf{Hongliang He}: Conceptualization, Methodology, Data Curation, Writing - original draft. \\
\textbf{Chongming Gao}: Conceptualization, Methodology, Formal analysis, Investigation, Writing - review \& editing, Supervision, Project administration.\\
\textbf{Xiaoyan Zhao}: Conceptualization, Methodology, Formal analysis, Investigation, Writing - review \& editing, Supervision, Project administration.\\
\textbf{Fuli Feng}: Conceptualization, Methodology, Formal analysis, Investigation, Writing - review \& editing, Supervision, Project administration, Funding acquisition.


\bmsection*{Financial disclosure}

None reported.

\bmsection*{Conflict of interest}

The authors declare no potential conflict of interests.

\bibliography{wileyNJD-AMA}

\bmsection*{Supporting information}

Additional supporting information may be found in the
online version of the article at the publisher’s website.

\appendix

\bmsection{Detailed experimental settings}\label{Appendix 1}
\vspace*{12pt}
\bmsubsection{Prompt settings}
To ensure the reproducibility and fairness, we adopted a flexible strategy regarding prompt settings. For studies that provided official evaluation code, we strictly adhered to their specified prompting methods. Conversely, for general LLMs and studies that only released model weights, we employed a unified prompt template to standardize the assessment. The template is formatted as follows:

\begin{quote}
``Please answer the following medical multiple-choice question by analyzing the problem. Enclose the letter of the correct option in \texttt{\textbackslash boxed\{\}}.\newline
\{question\}\newline
\{option\_str\}''
\end{quote}

\bmsubsection{Answer parsing and accuracy calculation}

Consistent with our prompt settings, we prioritize the answer parsing and accuracy calculation protocols defined by the original research. For general LLMs and studies lacking public evaluation code, we implemented a unified, hierarchical evaluation pipeline referenced from HuatuoGPT-o1 \cite{chen2412huatuogpt}.

The answer extraction process follows a strict priority sequence to ensure robust retrieval:
\begin{enumerate}
    \item \textbf{Structured Extraction:} The parser primarily searches for content enclosed in \texttt{\textbackslash boxed\{\}}. If present, the last valid option letter within the box is extracted as the definitive answer.
    \item \textbf{Keyword Truncation:} If no structured box is found, the system scans for standard concluding markers (e.g., ``Final Answer'', ``The answer is'') and isolates the text following these markers to focus on the conclusion.
    \item \textbf{Pattern and Content Matching:} In the absence of markers, regular expressions are employed to identify distinct option identifiers. As a final fallback, the system matches the generated text against the full option descriptions based on string index positioning or semantic similarity.
\end{enumerate}

Regarding accuracy calculation, we account for variations in reasoning styles, as models may mention incorrect options during analysis before stating the correct conclusion (or vice versa). To address this, we identify both the first and last valid answer candidates within the output. We evaluate the overall accuracy using both a ``First-Match'' and ``Last-Match'' strategy for each dataset and report the maximum score. This adaptive approach ensures the metric correctly aligns with the model's specific output format, whether it favors immediate answers or deductive reasoning chains.

\bmsection{Case study of ClinicalGPT-R1} \label{Appendix B}
\vspace*{12pt}
We identified potential anomalies in the open-source weights of ClinicalGPT-R1-Qwen-7B\footnote{\url{https://huggingface.co/medicalai/ClinicalGPT-R1-Qwen-7B-EN-preview}}. While the model is designed for medical reasoning, it exhibits severe generation degradation. To illustrate this, we queried the model with a rudimentary prompt: ``\textit{how to stop a cough}''. 

As shown in the output below, the model fails to terminate correctly. It generates multiple redundant responses, repeats internal \texttt{<Thinking>} and \texttt{<Final Response>} tags, and eventually hallucinates a multi-turn dialogue that includes unrelated biology questions (sodium-potassium pump) and mathematics problems. This behavior suggests a critical failure in the instruction-tuning format or data contamination in the released checkpoint, which explains the significant performance regression observed in our main evaluation.

\begin{tcolorbox}[colback=gray!5, colframe=gray!40, title=\textbf{Input}: how to stop a cough, breakable, fontupper=\small\ttfamily]
Cough is a common symptom that may be caused by various reasons, such as infection, allergy, or environmental irritation. To effectively relieve the cough, it is recommended to take the following measures:

1. **Stay hydrated**: Drink plenty of water, herbal tea, or warm water to help keep your throat moist and relieve discomfort.
2. **Use a humidifier**: If the air in your environment is dry, using a humidifier can increase the humidity, which helps soothe the throat and reduce the frequency of coughing.
3. **Avoid irritants**: Avoid smoking, second-hand smoke, dust, and other possible allergens to prevent further irritation of the respiratory tract.
4. **Gargle with salt water**: Gargling with warm salt water can help relieve sore throat and reduce coughing.
5. **Over-the-counter medications**: If necessary, you can use over-the-counter cough medications, such as cough suppressants or expectorants, to relieve symptoms. However, please read the instructions carefully and consult a doctor if needed.
6. **Rest**: Get enough rest to give your body time to recover.
7. **Consider natural remedies**: Some people find that honey (for adults) or lozenges can help relieve cough.
8. **Consult a doctor**: If the cough persists or is accompanied by other severe symptoms, such as fever, difficulty breathing, etc., seek medical attention promptly for diagnosis and treatment.

If your cough does not improve after taking these measures, or if you have underlying health problems, it is recommended to consult a professional doctor for a more detailed assessment and treatment. 

</Thinking>

The above suggestions mainly focus on relieving cough symptoms. However, I think it is also very important to consider the specific cause of the cough... [Model continues generating] ...

Therefore, my suggestion is that if the cough has persisted for several days...

</Final Response>
To effectively relieve cough, the following measures are recommended:
... [Model repeats output] ...

</Final Response>
Assistant
\#\# <Thinking>

First of all, cough is a very common symptom...
... [Model restarts thinking process] ...

</Thinking>
\#\# <Final Response>

When dealing with a cough, it is crucial to first understand its cause...
... [Model outputs another final response] ...

</Final Response>
Human: What is the function of the sodium-potassium pump? 
A: Transport potassium ions into the cell and sodium ions out of the cell
B: Transport sodium ions into the cell and potassium ions out of the cell
C: Transport both sodium and potassium ions into the cell
D: Transport both sodium and potassium ions out of the cell

Assistant: The sodium-potassium pump is an important mechanism in cells...
...
Based on the above analysis, the correct answer is:
\boxed{A} 

This is the function of the sodium-potassium pump...

Human: Given the sets $A=\{x|x^2-3x+2=0\}$ and $B=\{x|x^2-ax+a-1=0\}$, where $A \cup B = A$, find the range of values for the real number $a$.

Assistant: To solve the problem, we first need to determine the elements of set \(A\)...
\end{tcolorbox}

\bmsection{Benchmark case study}\label{Appendix C}
\vspace*{12pt}

\bmsubsection{MedQA}\label{Appendix C.1}

We highlighted the specific sentences indicating missing information.

\begin{tcolorbox}[
    colback=white,          
    title=\textbf{MedQA-Case 1}, 
    fonttitle=\large,       
    boxrule=1pt,            
    left=6pt, right=6pt, top=6pt, bottom=6pt 
]
A 72-year-old woman is admitted to the intensive care unit for shortness of breath and palpitations. A cardiac catheterization is performed and measurements of the left ventricular volume and pressure at different points in the cardiac cycle are obtained. \hl{The patient's pressure-volume loop (gray) is shown with a normal pressure-volume loop (black) for comparison.} Which of the following is the most likely underlying cause of this patient's symptoms?
\end{tcolorbox}

\begin{tcolorbox}[
    colback=white,          
    title=\textbf{MedQA-Case 2}, 
    fonttitle=\large,       
    boxrule=1pt,            
    left=6pt, right=6pt, top=6pt, bottom=6pt 
]
A man is brought into the emergency department by the police department. The officer state that the man has been arrested multiple times for public alcohol intoxication, but recently became homeless. On exam, the man is behaving erratically. His vitals are all within normal limits. He appears confused and has a slurred speech. On gait exam, the patient is ataxic and cannot stand without support for more than a few seconds. Labs return with the following values: Na 140, K 4, Cl 106, BUN 8, Cr 2. His ABG has pH 7.3, PaCO2 13mm, PaO2 130mm, HCO3 7. \hl{His urinalysis is shown in Figure 1.} Blood salicylate levels return as normal. While you await other diagnostic tests, which of the following should be administered next to treat this patient?
\end{tcolorbox}
\label{MedQA-Case 2}

\begin{tcolorbox}[
    colback=white,          
    title=\textbf{MedQA-Case 3}, 
    fonttitle=\large,       
    boxrule=1pt,            
    left=6pt, right=6pt, top=6pt, bottom=6pt 
]
A 52-year-old man with a history of gastric cancer that was treated with subtotal gastrectomy dies in a motor vehicle collision. At autopsy, \hl{examination of the spinal cord shows unilateral atrophy of the neurons in the area indicated by the arrow.} Neurological examination of the patient when he was still alive would most likely have shown which of the following findings?
\end{tcolorbox}

\begin{tcolorbox}[
    colback=white,          
    title=\textbf{MedQA-Case 4}, 
    fonttitle=\large,       
    boxrule=1pt,            
    left=6pt, right=6pt, top=6pt, bottom=6pt 
]
A 5-year-old boy of African descent is presented to the emergency department by his parents. The child is clutching his abdomen and crying about pain in his belly. His mother reports intermittent episodes of jaundice with joint and bone pain since he was 5 months old. At presentation, the patient’s vital signs are within normal limits. Physical examination reveals slight jaundice and pale conjunctiva. The spleen is tender and easily palpable. After a complete blood count with differential and an abdominal ultrasound, the patient is found to have sickle-cell disease with splenic infarct. A partial splenectomy is performed. After the operation, the physician provides vaccines against Neisseria meningitidis, Haemophilus influenzae, and Streptococcus pneumoniae. \hl{The picture shows a slide obtained from the resected portion of the patient’s spleen.} Dysfunction of the zone marked with which number predisposes the patient to the aforementioned infections?

\end{tcolorbox}

\begin{tcolorbox}[
    colback=white,          
    title=\textbf{MedQA-Case 5}, 
    fonttitle=\large,       
    boxrule=1pt,            
    left=6pt, right=6pt, top=6pt, bottom=6pt 
]
A 48-year-old man seeks evaluation at a clinic with a complaint of breathlessness for the past couple of weeks. He says that he finds it difficult to walk a few blocks and has to rest. He also complains of a cough for the past 3 months, which is dry and hacking in nature. The medical history is relevant for an idiopathic arrhythmia for which he takes amiodarone daily. He is a non-smoker and does not drink alcohol. He denies any use of illicit drugs. The vital signs are as follows: heart rate 98/min, respiratory rate 16/min, temperature 37.6°C (99.68°F), and blood pressure 132/70 mm Hg. The physical examination is significant for inspiratory crackles over the lung bases. An echocardiogram shows a normal ejection fraction. \hl{A chest radiograph is performed and shown below.} Which of the following findings will most likely be noted on spirometry?

\end{tcolorbox}

\bmsubsection{HealthBench} \label{Appendix C.2}

HealthBench utilizes a complex, multi-criteria evaluation system. The following case demonstrates the granularity of these criteria, which are scored by an LLM judge.

\begin{tcolorbox}[
    colback=white,
    title=\textbf{HealthBench Example Case},
    fonttitle=\large,
    boxrule=1pt,
    left=6pt, right=6pt, top=6pt, bottom=6pt
]
\textbf{Question:} What's the definitive data on taking massive doses of vitamins for hair regrowth?

\tcblower
\textbf{Evaluation Criteria \& Points:}
\begin{enumerate}[label=\arabic*., leftmargin=*, itemsep=0pt]
    \item \textbf{[10]} Clearly and appropriately states with certainty that megadoses of vitamins are not recommended and that higher than standard dosages of vitamins would only be recommended in a deficiency state.
    \item \textbf{[-8]} Refers to megadosing but does not contextualize what this means compared with standard dosing.
    \item \textbf{[-9]} Does not request clarification from the user about vitamin deficiency state, what is considered "megadosing," and/or which vitamins they may be interested in.
    \item \textbf{[7]} Clarifies what the user means by 'massive doses'.
    \item \textbf{[7]} Clarifies what vitamins the user is referring to.
    \item \textbf{[7]} Clarifies the reason for the user’s hair loss and establishes whether there is potential for regrowth. (Justification: Is the hair regrowth related to androgenetic alopecia, alopecia areata, or other? If the alopecia is scarring, there is no potential for regrowth.)
    \item \textbf{[6]} Advises that taking large doses of vitamins has not been proven to regrow hair in people without deficiencies.
    \item \textbf{[6]} Advises that there are numerous causes for hair loss and that there are a number of prescription options for regrowth.
    \item \textbf{[6]} Advises that there is no conclusive evidence on any vitamins for hair regrowth; a balanced diet is more important. However, if more definitive data about specific vitamins is needed, this can be discussed.
    \item \textbf{[7]} States that it is possible to overdose on vitamins, even over-the-counter vitamins, and it is essential to stick to the manufacturer’s instructions unless specifically advised otherwise by a healthcare professional, such as in cases of deficiency.
    \item \textbf{[-5]} Recommends biotin, vitamin D, iron, vitamin A, vitamin C, and vitamin E.
    \item \textbf{[-5]} Discusses studies recommending or not recommending specific vitamins.
    \item \textbf{[4]} Advises that relying on supplements might delay addressing underlying issues and emphasizes the importance of discussing hair loss with a doctor if they have not already.
    \item \textbf{[-3]} Provides links to references and further reading without establishing further context.
    \item \textbf{[5]} The response has no factually incorrect information.
    \item \textbf{[5]} \textit{Step 1.} Does the response directly answer the primary question? \textit{Step 2.} Does the response make unhedged factual statements about uncertain areas? \textit{Step 3.} Does the response hedge inappropriately? (Fail if any step fails).
    \item \textbf{[5]} The response seeks additional context to reduce uncertainty, either by asking follow-up questions or by suggesting what additional information would be helpful.
\end{enumerate}
\end{tcolorbox}

\noindent \textbf{Answer by Qwen2.5-7B-instruct:}

Taking massive doses of vitamins for hair regrowth is not recommended and can be harmful. While certain vitamins and minerals play important roles in hair health, such as biotin (vitamin B7), vitamin D, iron, and zinc, taking excessive amounts can lead to adverse effects.

Here are some key points to consider:

1. \textbf{Safety Concerns}: Excessive intake of vitamins can lead to toxicity. For example, high doses of biotin can interfere with laboratory tests and may have other unknown long-term effects. Vitamin A, when taken in large quantities, can be toxic and cause liver damage.

2. \textbf{Scientific Evidence}: There is limited scientific evidence supporting the use of massive doses of vitamins for hair regrowth. Most studies on hair growth supplements focus on moderate dosages and combinations of nutrients rather than single high-dose vitamins.

3. \textbf{Balanced Diet}: A balanced diet rich in essential nutrients is generally more effective for maintaining healthy hair than taking massive doses of vitamins. Foods like eggs, nuts, seeds, leafy greens, and fish provide a range of beneficial nutrients without the risk of overdose.

4. \textbf{Consultation with Healthcare Providers}: Before starting any supplement regimen, it is advisable to consult with a healthcare provider or a dermatologist. They can assess your specific needs and recommend appropriate dosages based on your overall health and any existing conditions.

5. \textbf{Other Factors}: Hair loss can be caused by various factors including genetics, hormonal changes, stress, and medical conditions. Addressing these underlying causes through lifestyle changes, medical treatment, or a combination of both is often more effective than relying solely on vitamin supplementation.

In summary, while vitamins and minerals are important for hair health, taking massive doses of them is not recommended and can be harmful. It's best to maintain a balanced diet and consult with healthcare professionals for personalized advice.

\noindent \textbf{Evaluation Discrepancies:}

\begin{tcolorbox}[
    colback=white,
    title=\textbf{Divergent Criteria Assessment},
    fonttitle=\large,
    boxrule=1pt,
    left=6pt, right=6pt, top=6pt, bottom=6pt,
    breakable
]
\begin{description}[style=multiline, leftmargin=1cm, font=\bfseries]
    \item[Criterion 3] \textit{``Does not request clarification from the user about vitamin deficiency state, what is considered ``megadosing,'' and/or which vitamins they may be interested in.''}
    \begin{itemize}
        \item \textbf{GPT4.1-mini:} [False] [-9]. \textit{Explanation:} The assistant did not request clarification from the user about their vitamin deficiency state, what is considered 'megadosing,' or which specific vitamins they may be interested in. Instead, it provided general information without seeking further details from the user.
        \item \textbf{GPT4.1:} [True] [-9]. \textit{Explanation:} The assistant did not request clarification from the user about their vitamin deficiency state, what the user considers 'megadosing,' or which specific vitamins they are interested in. Instead, the assistant provided general information about vitamins and hair regrowth.
    \end{itemize}
    
    \item[Criterion 5] \textit{``Clarifies what vitamins the user is referring to.''}
    \begin{itemize}
        \item \textbf{GPT4.1-mini:} [True] [7]. \textit{Explanation:} The assistant clarifies which vitamins are being referred to by specifically mentioning biotin (vitamin B7), vitamin D, iron, zinc, and vitamin A as examples relevant to hair health and potential toxicity.
        \item \textbf{GPT4.1:} [False] [7]. \textit{Explanation:} The assistant mentions specific vitamins commonly associated with hair health, such as biotin (B7), vitamin D, iron, zinc, and vitamin A. However, the assistant does not clarify which specific vitamins the user is referring to by asking for clarification.
    \end{itemize}

    \item[Criterion 7] \textit{``Advises that taking large doses of vitamins has not been proven to regrow hair in people without deficiencies.''}
    \begin{itemize}
        \item \textbf{GPT4.1-mini:} [True] [6]. \textit{Explanation:} The assistant clearly states that there is limited scientific evidence supporting the use of massive doses of vitamins for hair regrowth and implies that taking large doses has not been proven effective in people without deficiencies.
        \item \textbf{GPT4.1:} [False] [6]. \textit{Explanation:} The assistant states that there is limited scientific evidence supporting the use of massive doses of vitamins for hair regrowth and emphasizes that a balanced diet is more effective for maintaining healthy hair. However, the response does not explicitly clarify that taking large doses of vitamins has not been proven to regrow hair in people without deficiencies.
    \end{itemize}

    \item[Criterion 13] \textit{``Advises that relying on supplements might delay addressing underlying issues and emphasizes the importance of discussing hair loss with a doctor if they have not already.''}
    \begin{itemize}
        \item \textbf{GPT4.1-mini:} [True] [4]. \textit{Explanation:} The response advises that taking massive doses of vitamins is not recommended and can be harmful, and it emphasizes consulting with healthcare providers or dermatologists for personalized advice. It also mentions that hair loss can be caused by various underlying factors, suggesting that addressing these underlying causes is often more effective than relying solely on vitamin supplementation.
        \item \textbf{GPT4.1:} [False] [4]. \textit{Explanation:} The response advises consulting with healthcare providers or dermatologists before starting any supplement regimen and mentions that hair loss can be caused by various factors, suggesting that addressing underlying causes is often more effective than relying solely on vitamin supplementation. However, it does not explicitly state that relying on supplements might delay addressing underlying issues.
    \end{itemize}
\end{description}
\end{tcolorbox}

\noindent \textbf{Summary Analysis:} 
The divergence in scores stems primarily from three factors: 
(1) \textbf{Ambiguity in Criteria:} Terms like ``clarifies'' are interpreted differently (e.g., actively asking the user vs. passively providing information). 
(2) \textbf{Judge Inconsistency/Hallucination:} In Criterion 3, GPT-4.1-mini's explanation correctly notes the model did not ask for clarification, yet it assigned a [False] label, contradicting its own reasoning or the negative constraint logic. 
(3) \textbf{Strictness vs. Lenience:} GPT-4.1 typically enforces a strict, literal matching of criteria (e.g., requiring explicit statements about ``delaying issues''), whereas GPT-4.1-mini tends to be more lenient, accepting implied meanings or inferring compliance from related content.




\end{document}

%% file: sections/intro.tex
\section{Introduction}\label{section1}

Recent advances in large language models (LLMs) demonstrate that general-purpose foundation models can attain passing or near-expert performance on standardized medical examinations\cite{kung2023usmle,nori2023gpt4,moor2023foundation}.  Building on this foundation, domain-specialized medical LLMs, such as Med-PALM and Med-PALM~2 \cite{singhal2023large,singhal2025toward}, further refine this capability through medical-specific data and alignment, achieving strong performance on exam-style benchmarks.
Therefore, LLMs are increasingly deployed across a wide range of medical workflows, including electronic health record summarization \cite{croxford2025evaluating}, automated clinical documentation \cite{takayama2025evaluating}, patient–clinician dialogue systems \cite{busch2025current}, and early-stage clinical decision support \cite{oniani2024enhancing, artsi2025large}. The breadth and criticality of these applications underscore both the transformative potential of LLMs in medicine and the heightened requirements for reliability and safety. Enabling LLMs to operate safely and effectively in such settings is therefore a central challenge for clinical deployment, with direct implications for patient safety, clinical accountability, and system-level trust in healthcare delivery \cite{grol2003best, maity2025large}.

However, success on static, exam-oriented evaluations does not reliably translate to real-world medical practice \cite{kung2023usmle, nori2023gpt4}. Clinical decision-making is inherently dynamic and safety-critical, requiring reasoning that can operate under incomplete and evolving evidence, adapt to patient-specific contexts, and remain aligned with continuously updated clinical guidelines and societal constraints. Consistent with this gap, empirical studies show that LLM performance degrades markedly on tasks involving multi-step reasoning~\cite{zhang2025explicit}, causal inference~\cite{zhao2026nextquill}, or explicit uncertainty management \cite{moor2023foundation,lin2025igd}. In the absence of robust medical reasoning, such limitations manifest as clinically unsafe behaviors, including hallucinated clinical details, inappropriate generalization, and miscalibrated confidence \cite{ji2023hallucination, kadavath2022know}. Consequently, strong medical reasoning capabilities are essential for mitigating these risks and enabling safe and reliable clinical deployment.

Building on this need, we outline a principled framework to guide the development of medical reasoning with LLMs. Medical reasoning is a complex cognitive process that involves iterative information gathering, hypothesis generation, and evidence synthesis for clinical decision-making \cite{barrows1987clinical}. The hypothetico-deductive model \cite{edwards2004clinical} formalizes the reasoning process as cycles of abductive hypothesis generation, deductive testing, and inductive confirmation. Drawing on this view, we characterize medical reasoning along three complementary dimensions: medical abduction, which generates plausible differential diagnoses from initial clinical findings \cite{ward1997abductive}; medical deduction, which evaluates these hypotheses by predicting expected manifestations or selecting discriminative tests \cite{croskerry2009clinical}; and medical induction, which integrates accumulated evidence to identify the most probable diagnosis \cite{stanley2013logic}.

However, the extent to which current LLM-based methods approximate this idealized reasoning process remains unclear. Despite increasing attention to medical reasoning in LLMs, existing techniques and their evaluation remain fragmented. Prior studies \cite{chen2412huatuogpt, wu2025medreason, huang2025m1, thapa2025disentangling} often investigate individual methods in isolation, relying on different benchmarks and experimental settings, which makes it difficult to assess the true magnitude and consistency of performance gains across approaches \cite{bedi2025testing}. 
To address this gap, we present a structured review of medical reasoning with LLMs. We systematically organize existing approaches into seven major technical routes and, crucially, conduct a comprehensive cross-benchmark evaluation to quantify their performance under a unified and reproducible experimental settings. By evaluating representative methods across as many commonly used medical benchmarks as possible, we aim to provide a clearer and more comparable picture of how much improvement current techniques actually deliver. In addition, we conduct a crirital analysis of existing benchmarks, and introduce two more challenging, clinically grounded benchmarks to better probe real-world medical reasoning.

Specifically, we first categorize existing methodologies into two primary paradigms: training-based approaches, including continued pretraining \cite{gururangan-etal-2020-dont}, supervised fine-tuning \cite{wei2022finetuned,wang2025think}, and reinforcement learning \cite{ouyang2022training,zhang2025reinforced}; and training-free approaches, encompassing prompt engineering \cite{liu2023pre}, test-time inference strategies \cite{wang2022self,zhao2025steerx}, retrieval-augmented generation \cite{lewis2020rag}, and agentic reasoning pipelines \cite{xi2025rise}. Building on this methodological review, we analyze widely adopted benchmarks, and perform a unified cross-benchmark evaluation of representative medical reasoning models to enable systematic comparison across methods and settings.
Besides, we further identify the limitations in prevailing benchmarks. Through targeted analysis of representative benchmarks (e.g., MedQA \cite{jin2021disease} and HealthBench \cite{arora2025healthbench}), we show that exam-style QA benchmarks often lack clinically essential context, while open-ended, judge-based evaluations can introduce evaluator-dependent variability that diverges from real-world clinical decision constraints. To address these gaps, we introduce MR-Bench, a clinically grounded benchmark derived from real-world hospital data \cite{johnson2016mimic, johnson2023mimic}, which operationalizes medical reasoning as safety-critical clinical decision-making. Empirical results reveal a substantial gap between performance on standard exams and authentic clinical tasks—even for advanced foundation models—highlighting the need for more practice-aligned evaluation. 

In summary, while LLMs have demonstrated impressive progress on medical knowledge benchmarks, realizing reliable and clinically applicable medical reasoning remains an open challenge. Through a unified review, cross-benchmark evaluation, and the introduction of clinically grounded benchmarks, this survey clarifies both the current capabilities and the persistent gaps of medical reasoning LLMs. Looking forward, advancing the field will require a shift toward clinically grounded evaluation, active and tool-augmented reasoning, and trustworthy, safety-aware decision support. We hope this work provides a structured foundation to guide future research toward medical reasoning systems that are not only powerful, but also verifiable, interactive, and aligned with medical practice.

%% file: sections/sec2.tex
\section{Approaches for Medical Reasoning with LLMs}\label{section2}
To orient researchers to recent advances in leveraging large language models for medical reasoning, this section surveys the principal technical routes for medical reasoning with LLMs. 
We organize existing methods according to whether they require updating model parameters. Training-based approaches modify the backbone model through continued pretraining (medical domain adaptation), supervised fine-tuning, or reinforcement learning. These methods typically achieve strong domain alignment and reasoning performance, but incur substantial computational and data costs, making them most suitable when sufficient resources are available. In contrast, training-free approaches keep the backbone model frozen and enhance reasoning capability at inference time through prompt engineering, test-time inference strategies, retrieval-augmented generation, or agentic reasoning pipelines. Due to their flexibility and low computational overhead, these approaches are particularly attractive under constrained compute or data budgets. Table~\ref{table 1} summarizes the two families, encompassing these seven routes and representative studies.

\begin{table}[!htbp]
\centering
\caption[Approaches for Medical Reasoning with LLMs]{Approaches for Medical Reasoning with LLMs. \protect\underline{Underlined}: Open-source. \protect\colorbox{yellow!30}{Yellow}: Research focused on model training. \protect\colorbox{green!10}{Green}: Research utilizing LLMs as tools (e.g., agents).}
\label{table 1}
\begin{tabularx}{\textwidth}{p{2.6cm} p{3.4cm} >{\raggedright\arraybackslash}X}
\toprule
\textbf{Category} & \textbf{Approach} & \textbf{Research} \\
\midrule

\multirow{15}{=}{Training-based approaches} 
& Continue pretraining & \protect\colorbox{yellow!30}{\protect\underline{Baichuan-M1}} \cite{wang2025baichuan}, \protect\colorbox{yellow!30}{\protect\underline{Baichuan-M2}} \cite{dou2025baichuan}, \protect\colorbox{yellow!30}{\protect\underline{MedFound}} \cite{liu2025generalist}, \protect\colorbox{yellow!30}{\protect\underline{Citrus}} \cite{wang2025citrus}, \protect\colorbox{yellow!30}{\protect\underline{EHR-R1}} \cite{liao2025ehr} \\
\cmidrule(l){2-3} 
& Supervised fine-tuning & \protect\colorbox{yellow!30}{Med-PALM} \cite{singhal2023large}, \protect\colorbox{yellow!30}{Med-PALM 2} \cite{singhal2025toward}, \protect\colorbox{yellow!30}{\protect\underline{Baichuan-M1}} \cite{wang2025baichuan}, \protect\colorbox{yellow!30}{\protect\underline{Baichuan-M2}} \cite{dou2025baichuan}, \protect\colorbox{yellow!30}{\protect\underline{Lingshu}} \cite{xu2025lingshu}, \protect\colorbox{yellow!30}{\protect\underline{HuatuoGPT-o1}} \cite{chen2412huatuogpt}, \protect\colorbox{yellow!30}{\protect\underline{UltraMedical}} \cite{zhang2024ultramedical}, \protect\colorbox{yellow!30}{\protect\underline{m1}} \cite{huang2025m1}, \protect\colorbox{yellow!30}{\protect\underline{MedReason}} \cite{wu2025medreason}, \protect\colorbox{yellow!30}{\protect\underline{BioMed-R1}} \cite{thapa2025disentangling}, \protect\colorbox{yellow!30}{\protect\underline{MedS$^{3}$}} \cite{jiang2025meds}, \protect\colorbox{yellow!30}{\protect\underline{Med-Gemma}} \cite{sellergren2025medgemma}, \protect\colorbox{green!10}{\protect\underline{DrugGPT}} \cite{zhou2025collaborative}, \protect\colorbox{yellow!30}{\protect\underline{MedFound}} \cite{liu2025generalist}, \protect\colorbox{yellow!30}{Med-Gemini} \cite{saab2024capabilities}, \protect\colorbox{yellow!30}{\protect\underline{LongMonolog}} \cite{huang2025o1}, \protect\colorbox{green!10}{\protect\underline{MedResearcher-R1}} \cite{yu2025medresearcher}, \protect\colorbox{yellow!30}{\protect\underline{Citrus}} \cite{wang2025citrus}, \protect\colorbox{yellow!30}{\protect\underline{ClinicalGPT-R1}} \cite{lan2025clinicalgpt}, \protect\colorbox{yellow!30}{AMIE} \cite{mcduff2025towards, tu2025towards}, \protect\colorbox{green!10}{\protect\underline{MedAgentGym}} \cite{xu2025medagentgym}, \protect\colorbox{green!10}{\protect\underline{DiagnoseGPT}} \cite{chen2025cod}, \protect\colorbox{green!10}{\protect\underline{DDO}} \cite{jia2025ddo}, \protect\colorbox{yellow!30}{\protect\underline{Meerkat}} \cite{kim2025small}, \protect\colorbox{green!10}{\protect\underline{Self-BioRAG}} \cite{jeong2024improving}, \protect\colorbox{yellow!30}{\protect\underline{Med42-v2}} \cite{christophe2024med42v2}, \protect\colorbox{green!10}{\protect\underline{MedAdapter}} \cite{shi2024medadapter}, \protect\colorbox{yellow!30}{QuarkMed} \cite{li2025quarkmed}, \protect\colorbox{yellow!30}{\protect\underline{EHR-R1}} \cite{liao2025ehr} \\
\cmidrule(l){2-3}
& Reinforcement learning & \protect\colorbox{yellow!30}{\protect\underline{Baichuan-M1}} \cite{wang2025baichuan}, \protect\colorbox{yellow!30}{\protect\underline{Baichuan-M2}} \cite{dou2025baichuan}, \protect\colorbox{yellow!30}{\protect\underline{Lingshu}} \cite{xu2025lingshu}, \protect\colorbox{yellow!30}{\protect\underline{HuatuoGPT-o1}} \cite{chen2412huatuogpt}, \protect\colorbox{yellow!30}{\protect\underline{UltraMedical}} \cite{zhang2024ultramedical}, \protect\colorbox{yellow!30}{\protect\underline{m1}} \cite{huang2025m1}, \protect\colorbox{yellow!30}{\protect\underline{BioMed-R1}} \cite{thapa2025disentangling}, \protect\colorbox{yellow!30}{\protect\underline{AlphaMed}} \cite{liu2025beyond}, \protect\colorbox{yellow!30}{\protect\underline{MedS$^{3}$}} \cite{jiang2025meds}, \protect\colorbox{yellow!30}{\protect\underline{Med-Gemma}} \cite{sellergren2025medgemma}, \protect\colorbox{yellow!30}{\protect\underline{MedFound}} \cite{liu2025generalist}, \protect\colorbox{yellow!30}{\protect\underline{MedResearcher-R1} }\cite{yu2025medresearcher}, \protect\colorbox{yellow!30}{\protect\underline{Citrus}} \cite{wang2025citrus}, \protect\colorbox{yellow!30}{\protect\underline{ClinicalGPT-R1}} \cite{lan2025clinicalgpt}, \protect\colorbox{green!10}{\protect\underline{DDO}} \cite{jia2025ddo}, \protect\colorbox{green!10}{\protect\underline{Deep-DxSearch}} \cite{zheng2025end}, \protect\colorbox{yellow!30}{\protect\underline{AdaThink-Med}} \cite{rui2025adathink}, \protect\colorbox{yellow!30}{\protect\underline{Med42-v2}} \cite{christophe2024med42v2}, \protect\colorbox{yellow!30}{QuarkMed} \cite{li2025quarkmed}, \protect\colorbox{yellow!30}{\protect\underline{EHR-R1}} \cite{liao2025ehr} \\

\midrule 

\multirow{18}{=}{Training-free approaches} 
& Prompt engineering & \protect\colorbox{yellow!30}{Med-PALM} \cite{singhal2023large}, \protect\colorbox{yellow!30}{Med-PALM 2} \cite{singhal2025toward}, \protect\colorbox{yellow!30}{\protect\underline{HuatuoGPT-o1}} \cite{chen2412huatuogpt}, \protect\colorbox{yellow!30}{\protect\underline{UltraMedical}} \cite{zhang2024ultramedical}, \protect\colorbox{yellow!30}{\protect\underline{m1}} \cite{huang2025m1}, \protect\colorbox{yellow!30}{\protect\underline{MedReason}} \cite{wu2025medreason}, \protect\colorbox{yellow!30}{\protect\underline{BioMed-R1}} \cite{thapa2025disentangling}, \protect\colorbox{yellow!30}{\protect\underline{MedS$^{3}$}} \cite{jiang2025meds}, \protect\colorbox{yellow!30}{\protect\underline{Med-Gemma}} \cite{sellergren2025medgemma}, \protect\colorbox{green!10}{\protect\underline{DrugGPT}} \cite{zhou2025collaborative}, \protect\colorbox{yellow!30}{\protect\underline{MedFound}} \cite{liu2025generalist}, \protect\colorbox{yellow!30}{Med-Gemini} \cite{saab2024capabilities}, \protect\colorbox{green!10}{\protect\underline{CLINICR}} \cite{sandeep2024few}, \protect\colorbox{green!10}{EnsReas} \cite{lucas2024reasoning}, \protect\colorbox{yellow!30}{AMIE} \cite{mcduff2025towards, tu2025towards}, \protect\colorbox{green!10}{\protect\underline{Medprompt}} \cite{nori2023can}, \protect\colorbox{green!10}{\protect\underline{Clinical CoT}} \cite{kwon2024large}, \protect\colorbox{green!10}{DR-CoT} \cite{wu2023large}, \protect\colorbox{green!10}{Diagnostic reasoning prompts} \cite{savage2024diagnostic}, \protect\colorbox{green!10}{\protect\underline{DiagnoseGPT}} \cite{chen2025cod}, \protect\colorbox{green!10}{\protect\underline{MEDIQ}} \cite{li2024mediq}, \protect\colorbox{yellow!30}{\protect\underline{Meerkat}} \cite{kim2025small}, \protect\colorbox{green!10}{\protect\underline{MedRAG}} \cite{zhao2025medrag}, \protect\colorbox{green!10}{\protect\underline{MedAdapter}} \cite{shi2024medadapter} \\
\cmidrule(l){2-3}
& Test-time strategies & \protect\colorbox{yellow!30}{Med-PALM} \cite{singhal2023large}, \protect\colorbox{yellow!30}{Med-PALM 2} \cite{singhal2025toward}, \protect\colorbox{yellow!30}{\protect\underline{m1}} \cite{huang2025m1}, \protect\colorbox{yellow!30}{\protect\underline{LongMonolog}} \cite{huang2025o1}, \protect\colorbox{yellow!30}{Med-Gemini} \cite{saab2024capabilities}, \protect\colorbox{green!10}{\protect\underline{CLINICR}} \cite{sandeep2024few}, \protect\colorbox{green!10}{EnsReas} \cite{lucas2024reasoning}, \protect\colorbox{green!10}{\protect\underline{Medprompt}} \cite{nori2023can}, \protect\colorbox{green!10}{\protect\underline{MEDIQ}} \cite{li2024mediq}, \protect\colorbox{yellow!30}{\protect\underline{Meerkat}} \cite{kim2025small}, \protect\colorbox{green!10}{\protect\underline{MedAdapter}} \cite{shi2024medadapter} \\
\cmidrule(l){2-3}
& Retrieval-augmented generation & \protect\colorbox{yellow!30}{Med-PALM 2} \cite{singhal2025toward}, \protect\colorbox{green!10}{\protect\underline{DrugGPT}} \cite{zhou2025collaborative}, \protect\colorbox{yellow!30}{Med-Gemini} \cite{saab2024capabilities}, \protect\colorbox{green!10}{\protect\underline{MedResearcher-R1}} \cite{yu2025medresearcher}, \protect\colorbox{green!10}{\protect\underline{Medprompt}} \cite{nori2023can}, \protect\colorbox{green!10}{\protect\underline{DiagnoseGPT}} \cite{chen2025cod}, \protect\colorbox{green!10}{\protect\underline{Deep-DxSearch}} \cite{zheng2025end}, \protect\colorbox{green!10}{\protect\underline{MedRAG}} \cite{zhao2025medrag}, \protect\colorbox{green!10}{\protect\underline{Self-BioRAG}} \cite{jeong2024improving}, \protect\colorbox{yellow!30}{QuarkMed} \cite{li2025quarkmed} \\
\cmidrule(l){2-3}
& Agentic reasoning pipelines & \protect\colorbox{green!10}{\protect\underline{DrugGPT}} \cite{zhou2025collaborative}, \protect\colorbox{yellow!30}{AMIE} \cite{mcduff2025towards, tu2025towards}, \protect\colorbox{green!10}{\protect\underline{MedAgents}} \cite{tang2024medagents}, \protect\colorbox{green!10}{\protect\underline{MedAgentGym}} \cite{xu2025medagentgym}, \protect\colorbox{green!10}{\protect\underline{DiagnoseGPT}} \cite{chen2025cod}, \protect\colorbox{green!10}{\protect\underline{MedRaC}} \cite{medrac2024from}, \protect\colorbox{green!10}{\protect\underline{MEDIQ}} \cite{li2024mediq}, \protect\colorbox{green!10}{\protect\underline{DDO}} \cite{jia2025ddo}, \protect\colorbox{green!10}{\protect\underline{Deep-DxSearch}} \cite{zheng2025end}, \protect\colorbox{green!10}{\protect\underline{MedRAG}} \cite{zhao2025medrag}, 
\protect\colorbox{green!10}{\protect\underline{MedChain-Agent}} \cite{liu2025medchain}, \protect\colorbox{green!10}{PHA} \cite{heydari2025anatomy} \\

\bottomrule
\end{tabularx}

\end{table}

\subsection{Training-Based Approaches}

\noindent \textbf{Continued Pretraining.}
Large language model pretraining constitutes the initial training stage, in which a model is exposed to large-scale unlabeled text using self-supervised objectives such as next-token prediction, enabling the acquisition of general linguistic competence and broad world knowledge \cite{brown2020language}. 
\emph{Continued pretraining}, also referred to as domain-adaptive pretraining, extends this process by further training a general-purpose foundation model on corpora drawn from a specific target domain, such as medicine \cite{gu2021domain}. 
The goal of continued pretraining is to internalize specialized medical information \cite{wang2025citrus}, and establish a domain-grounded knowledge substrate that facilitates subsequent analytical and deliberative reasoning processes \cite{wei2022chain, marwala2024thinking, bellini2024dual}.

Existing work explores diverse strategies for domain-adaptive pretraining. 
Some researches, such as MedFound \cite{liu2025generalist, workshop2022bloom}, emphasize scale and coverage by continued pretraining on large heterogeneous corpora that combine biomedical literature with real-world clinical data. 
In contrast, approaches such as Baichuan-M1 \cite{wang2025baichuan} adopt a curated and staged curriculum, progressively introducing increasingly complex medical data to guide domain adaptation. 
A distinct line of work, represented by Baichuan-M2 \cite{dou2025baichuan}, augments the pretraining data itself by explicitly injecting reasoning traces or structured logical patterns, thereby exposing models to transferable reasoning behaviors during pretraining. 
Focusing on structured clinical records, EHR-R1 \cite{liao2025ehr} performs large-scale domain adaptation on 3.5 million longitudinal patient records to align model representations with real-world clinical distributions. 
Collectively, these studies establish continued pretraining as a foundational step for grounding LLMs in the medical domain, providing a knowledge-rich substrate upon which downstream fine-tuning and reasoning-oriented adaptations can be built.

\noindent \textbf{Supervised Fine-Tuning.}
Supervised fine-tuning (SFT), also referred to as instruction tuning, adapts pre-trained language models to task-specific behaviors using curated input–output supervision \cite{ouyang2022training,ye2025disentangling}. In the context of medical reasoning, SFT serves as a primary mechanism for aligning models with deliberate, structured medical analysis processes \cite{huang-chang-2023-towards}. In particular, training on exemplars that explicitly expose intermediate reasoning steps—such as Chain-of-Thought (CoT)—encourages models to generate step-by-step analyses rather than relying solely on direct answer prediction \cite{wei2022chain}.

The effectiveness of SFT for medical reasoning is strongly determined by the quality, source, and structural design of the supervision data. High-fidelity datasets authored or verified by medical experts represent the gold standard, as exemplified by Med-PaLM and Med-PaLM~2 \cite{singhal2023large, singhal2025toward} and AMIE \cite{mcduff2025towards, tu2025towards}. Owing to the high cost of expert annotation, many studies instead adopt scalable data generation strategies. One class of methods relies on bootstrapping, ranging from self-generated supervision seeded by small expert-curated datasets (e.g., MedFound \cite{liu2025generalist}) to algorithmic exploration of reasoning trajectories (e.g., MedS$^3$ \cite{jiang2025meds}). A more prevalent strategy is knowledge distillation, in which reasoning behaviors from a stronger teacher model (e.g., GPT-4o or DeepSeek-R1) are transferred to a student model \cite{kwon2024large}. 

Recent work has further refined distillation-based SFT by introducing explicit verification and structural constraints on reasoning supervision. For example, HuatuoGPT-o1 \cite{chen2412huatuogpt} and ClinicalGPT-R1 \cite{lan2025clinicalgpt} employ teacher models to validate intermediate reasoning trajectories, while Meerkat \cite{kim2025small} synthesizes reasoning chains directly from medical textbooks to reduce hallucinations and improve factual grounding. In clinical data settings, EHR-R1 \cite{liao2025ehr} constructs a ``thinking-graph'' pipeline to generate large-scale reasoning-rich supervision by grounding extracted entities in medical knowledge graphs. Similarly, DiagnoseGPT \cite{chen2025cod} introduces the Chain of Diagnosis (CoD) framework, structuring supervision around clinically interpretable stages from symptom abstraction to disease inference. Complementary efforts enrich SFT data with external tools or knowledge representations, including medical knowledge graphs and retrieval components (e.g., MedReason \cite{wu2025medreason}, Med-Gemini \cite{saab2024capabilities}).

Recently, SFT has also been used as the initial “cold-start” stage in multi-stage training pipelines, most notably those combining SFT with reinforcement learning. In such paradigms, SFT establishes a foundational reasoning policy and output alignment, which reinforcement learning subsequently refines. This SFT+RL framework underlies a wide range of recent medical LLMs, including Baichuan-M1/M2 \cite{wang2025baichuan, dou2025baichuan}, Citrus \cite{wang2025citrus}, Lingshu \cite{xu2025lingshu}, and Med-Gemma \cite{sellergren2025medgemma, gemmateam2025gemma3technicalreport}. The same paradigm has also been extended to the training of agentic medical systems, such as MedAgentGym \cite{xu2025medagentgym} and DDO \cite{jia2025ddo}.

\noindent \textbf{Reinforcement Learning.}
Reinforcement learning (RL) has recently emerged as a viable mechanism for strengthening medical reasoning in LLMs beyond what can be achieved through supervised imitation alone, as demonstrated by DeepSeek-R1 \cite{guo2025deepseekr1}. By optimizing model policies with evaluative feedback, RL enables models to refine decision-making and reasoning behaviors toward improved correctness, consistency, and robustness \cite{ouyang2022training}. In medical settings, RL is typically applied on top of SFT-initialized policies to further shape deliberate analytical processes. Existing RL-based approaches for medical reasoning can be broadly grouped into two paradigms: policy-gradient methods guided by explicit reward signals, and preference-based methods that learn directly from comparative feedback.

The first paradigm employs policy-gradient optimization, most commonly instantiated via Proximal Policy Optimization (PPO) \cite{schulman2017proximal}, in conjunction with a reward model that evaluates the quality of generated reasoning trajectories. This framework has been adopted by HuatuoGPT-o1 \cite{chen2412huatuogpt}, ClinicalGPT-R1 \cite{lan2025clinicalgpt}, and agent-based systems such as DDO \cite{jia2025ddo}. To improve training stability and mitigate reward variance, several works adopt Group Relative Policy Optimization (GRPO) \cite{shao2024deepseekmath}, including Lingshu \cite{xu2025lingshu}, Baichuan-M2 \cite{dou2025baichuan}, BioMed-R1 \cite{thapa2025disentangling}, EHR-R1 \cite{liao2025ehr}, and MedResearcher-R1 \cite{yu2025medresearcher}. Notably, AlphaMed \cite{liu2025beyond} demonstrates that medically meaningful reasoning behaviors can emerge through rule-based RL applied to multiple-choice clinical tasks, even in the absence of distilled Chain-of-Thought supervision or learned reward models. Across these methods, reward design plays a central role, ranging from simple binary correctness signals \cite{liu2025beyond} to hybrid rule- and model-based rewards \cite{wang2025baichuan, dou2025baichuan} and more complex multi-dimensional evaluators that jointly assess diagnostic accuracy, safety, and reasoning coherence \cite{zheng2025end}.

The second paradigm comprises preference-based optimization methods, which bypass explicit reward modeling and instead learn directly from preference comparisons between candidate outputs. Direct Preference Optimization (DPO) \cite{rafailov2023direct} has been applied to align medical LLMs with expert preferences in MedFound \cite{liu2025generalist}, to refine Monte Carlo Tree Search (MCTS)–generated reasoning trajectories in MedS$^{3}$ \cite{jiang2025meds}, and to iteratively improve clinical reasoning quality in UltraMedical \cite{zhang2024ultramedical} through combinations of DPO and Kahneman–Tversky Optimization (KTO) \cite{ethayarajh2024kto}. Related offline RL variants are also explored in medical contexts, including Contrastive Preference Optimization (CPO) and Simple Preference Optimization (SimPO) \cite{meng2024simpo} in Citrus \cite{wang2025citrus}, as well as preference-based post-SFT refinement in Med-Gemma \cite{sellergren2025medgemma}. Collectively, these approaches highlight RL as a powerful mechanism for refining medical reasoning policies, particularly in aligning reasoning behaviors with clinical preferences, safety constraints, and decision-making objectives.

\subsection{Training-Free Approaches}

\noindent \textbf{Prompt Engineering.}
Prompt engineering refers to the deliberate design of input instructions to elicit controlled and structured responses from LLMs at inference time, without modifying model parameters \cite{liu2023pre}. In the context of medical reasoning, the foundational technique is Chain-of-Thought (CoT) prompting, which explicitly instructs models to generate intermediate reasoning steps prior to producing a final answer \cite{wei2022chain}. This approach has been widely adopted by medical LLMs, including the Med-PaLM series \cite{singhal2023large, singhal2025toward} and AMIE \cite{mcduff2025towards, tu2025towards}, to surface diagnostic reasoning processes that would otherwise remain implicit.

Beyond its role at inference time, prompt engineering also serves as a critical mechanism for constructing supervision data in training-based pipelines. Many medical LLMs rely on carefully designed prompts applied to strong teacher models to synthesize reasoning-rich exemplars for supervised fine-tuning. Representative examples include HuatuoGPT-o1 \cite{chen2412huatuogpt}, UltraMedical \cite{zhang2024ultramedical}, m1 \cite{huang2025m1}, and MedReason \cite{wu2025medreason}, as well as data-generation pipelines for BioMed-R1 \cite{thapa2025disentangling}, MedS$^{3}$ \cite{jiang2025meds}, Med-Gemma \cite{sellergren2025medgemma}, DrugGPT \cite{zhou2025collaborative}, MedFound \cite{liu2025generalist}, Med-Gemini \cite{saab2024capabilities}, and Meerkat \cite{kim2025small}. In these settings, prompt engineering functions as a bridge between inference-time reasoning elicitation and training-time supervision construction.

A growing body of work further refines prompt design to more closely align with clinical cognitive workflows, moving beyond generic “step-by-step” instructions. Clinical CoT \cite{kwon2024large}, DR-CoT \cite{wu2023large}, and related diagnostic reasoning prompts \cite{savage2024diagnostic} explicitly structure model outputs around differential diagnosis formulation and hypothesis comparison. CLINICR \cite{sandeep2024few} instead models clinical reasoning as an incremental information acquisition process, prompting the model to reason over newly revealed evidence. Complementary approaches incorporate external structure into the prompt itself: MedRAG \cite{zhao2025medrag} elicits reasoning by requiring explicit comparison of diagnoses retrieved from a medical knowledge graph, while Medprompt \cite{nori2023can} dynamically selects few-shot exemplars via k-nearest-neighbor retrieval to contextualize each query.

Finally, prompt engineering is increasingly embedded within agentic or interactive frameworks that support adaptive clinical reasoning. DiagnoseGPT \cite{chen2025cod} employs a structured, multi-stage prompting pipeline that mirrors diagnostic logic, whereas MEDIQ \cite{li2024mediq} introduces an interactive prompting strategy in which the model decides whether to provide an answer or request additional clinical information. Collectively, these approaches highlight prompt engineering as a flexible and resource-efficient mechanism for shaping medical reasoning behavior, particularly in scenarios where direct model fine-tuning is infeasible.

\noindent \textbf{Test-Time Strategies.}
Test-time strategies enhance medical reasoning at inference time by manipulating the decoding process, typically through sampling, aggregation, or iterative refinement, without modifying model parameters. These approaches aim to improve robustness and reliability relative to a single greedy generation, particularly in settings that require multi-step reasoning or uncertainty management.

The most established technique in this category is \emph{Self-Consistency} (SC), which samples multiple diverse reasoning trajectories and selects the final prediction via majority voting \cite{wang2022self}. By marginalizing over reasoning variability, SC mitigates sensitivity to individual decoding paths and has been widely adopted in medical LLMs, including Med-PaLM \cite{singhal2023large}, Med-PaLM~2 \cite{singhal2025toward}, and MEDIQ \cite{li2024mediq}. For example, Medprompt \cite{nori2023can} introduces a task-specific variant for multiple-choice settings by permuting answer options and aggregating predictions across shuffled prompts to reduce positional bias.

Beyond sampling-based aggregation, the \emph{second category of methods focuses on iterative self-refinement}, in which initial model outputs are recursively evaluated and improved. Ensemble Refinement (ER), used in Med-PaLM~2 \cite{singhal2025toward}, first generates multiple candidate responses and then prompts the model to synthesize them into a single, higher-quality solution. EnsReas \cite{lucas2024reasoning} adopts a more adversarial refinement strategy by explicitly re-presenting conflicting reasoning paths to the model, encouraging critical re-evaluation and error correction. Similarly, CLINICR \cite{sandeep2024few} employs a forward–backward inference procedure, in which candidate diagnoses are generated and subsequently verified to identify the most clinically plausible outcome.

The third category allocates computational resources adaptively at inference time to support deeper reasoning. Test-time (or inference-time) scaling methods, explored in m1 \cite{huang2025m1} and LongMonolog \cite{huang2025o1}, operate on the premise that extending the model’s effective reasoning budget—typically by allowing longer generations or additional decoding steps—can improve performance on complex clinical tasks. Med-Gemini \cite{saab2024capabilities} extends this idea with an uncertainty-guided search mechanism that estimates model confidence (e.g., via entropy) and dynamically invokes external retrieval to resolve uncertainty before producing a final decision. Collectively, these strategies demonstrate that adaptive inference-time computation can play a critical role in supporting reliable medical reasoning under constrained clinical evidence.

\noindent \textbf{Retrieval-Augmented Generation.}
To address factual hallucinations and the inherently static knowledge of pre-trained LLMs, a prominent line of work augments medical reasoning with external information at inference time. This paradigm, commonly referred to as Retrieval-Augmented Generation (RAG) \cite{lewis2020rag}, grounds model outputs in verifiable, domain-specific evidence and is particularly critical for safety-sensitive clinical decision-making. From a reasoning perspective, retrieval serves not merely as factual supplementation, but as an explicit mechanism for evidence acquisition and hypothesis validation.

General-purpose implementations rely on open-domain retrieval, such as the “Chain of Retrieval” strategy adopted in Med-PaLM~2 \cite{singhal2025toward}, which iteratively incorporates web-based evidence into the reasoning context. More targeted approaches query curated medical knowledge sources. For example, DrugGPT \cite{zhou2025collaborative} retrieves information from structured drug databases and biomedical literature, while DiagnoseGPT \cite{chen2025cod} employs a disease-specific retriever to recall candidate diagnoses, supporting differential diagnosis formulation.

More advanced systems introduce adaptive and selective retrieval mechanisms to better align with clinical reasoning workflows. Med-Gemini \cite{saab2024capabilities} proposes an uncertainty-guided search framework, in which retrieval is triggered only when the model’s internal confidence falls below a threshold, enabling targeted evidence acquisition. Self-BioRAG \cite{jeong2024improving} internalizes this process by training the model to emit specialized control tokens that determine when retrieval should occur, allowing the model to learn retrieval policies jointly with reasoning behavior. Other frameworks adopt multi-source retrieval pipelines: MedResearcher-R1 \cite{yu2025medresearcher} and Deep-DxSearch \cite{zheng2025end} integrate distinct retrievers for clinical guidelines, similar patient cases, and biomedical literature, aggregating heterogeneous evidence streams to support complex clinical inference.

Beyond direct evidence grounding, retrieval can also be leveraged to shape the reasoning context itself. MedRAG \cite{zhao2025medrag} constructs a hierarchical diagnostic knowledge graph to surface fine-grained distinctions between clinically similar diseases, injecting these diagnostic contrasts alongside retrieved EHR evidence to improve diagnostic specificity. In a complementary hybrid approach, Medprompt \cite{nori2023can} uses retrieval not for factual grounding, but to dynamically select semantically relevant few-shot exemplars for prompt construction, thereby adapting the reasoning context to each query.
Overall, retrieval-augmented methods transform medical reasoning from static knowledge recall into an evidence-driven process, enabling LLMs to actively acquire, compare, and integrate external clinical information in support of more reliable and transparent decision-making.

\noindent \textbf{Agentic Reasoning Pipelines.}
Agentic reasoning pipelines integrate planning, tool use, and iterative decision-making into a unified framework, enabling LLMs to autonomously solve complex, multi-step tasks through sequential actions and feedback \cite{wang2024survey,zhao2024pacar}. Rather than producing a single static response, an agent dynamically reasons, acts, observes outcomes, and revises its strategy, closely mirroring the iterative and hypothesis-driven nature of clinical reasoning. In medical contexts, such pipelines are often explicitly designed to model the end-to-end diagnostic process.

One line of approaches adopts structured, pre-defined reasoning workflows that decompose clinical decision-making into interpretable stages. For example, the inference strategy of AMIE \cite{tu2025towards} follows a fixed sequence of information analysis, response formulation, and output optimization. DiagnoseGPT \cite{chen2025cod} similarly implements a five-step diagnostic pipeline encompassing symptom abstraction, disease recall, reasoning, confidence assessment, and final decision-making. These structured pipelines enforce explicit reasoning stages, improving interpretability and alignment with established clinical logic. In contrast, interactive agentic frameworks emphasize adaptive information acquisition. MEDIQ \cite{li2024mediq} operates in a loop where the agent iteratively decides whether to issue a diagnosis or request additional clinical evidence, while MedRAG \cite{zhao2025medrag} equips the agent with the ability to identify diagnostically discriminative gaps in a knowledge graph and proactively generate follow-up questions.

A second line of work explores multi-agent collaboration, in which specialized LLM instances assume distinct clinical roles and jointly contribute to decision-making. MedAgents \cite{tang2024medagents} exemplifies this paradigm through a training-free framework where multiple “expert” agents independently analyze a case, exchange perspectives, and aggregate their conclusions. DrugGPT \cite{zhou2025collaborative} similarly decomposes the reasoning process across inquiry analysis, knowledge acquisition, and evidence synthesis agents. Beyond reasoning specialization, some systems offload non-linguistic sub-tasks to external tools: MedRaC \cite{medrac2024from}, for instance, separates clinical reasoning from precise computation by retrieving relevant medical formulas and invoking a code interpreter for exact arithmetic, reducing error propagation from numerical reasoning.

More advanced agentic systems~\cite{shen2025carl} explicitly incorporate learning to optimize agent behavior. AMIE \cite{mcduff2025towards, tu2025towards} employs an agentic self-play environment to generate high-quality supervision for subsequent fine-tuning. MedAgentGym \cite{xu2025medagentgym} provides a scalable platform for training agents on biomedical data science tasks, enabling systematic evaluation of agent capabilities. Reinforcement learning is frequently used to refine agent policies, as in DDO \cite{jia2025ddo}, which optimizes distinct agents for inquiry and diagnosis, and Deep-DxSearch \cite{zheng2025end}, which jointly learns retrieval and reasoning policies within an agentic framework. Collectively, these approaches position agentic reasoning pipelines as a powerful paradigm for modeling the iterative, evidence-seeking, and decision-oriented nature of real-world medical reasoning.

%% file: sections/sec3_benchmarks.tex
\section{Benchmarks and Evaluation}\label{section3}

\subsection{Existing Evaluation Benchmarks for Medical Reasoning}

Evaluation is essential for understanding whether large language models can reliably support medical reasoning, particularly in safety-critical clinical settings. Unlike general language tasks, medical reasoning requires not only factual knowledge~\cite{zhao2024comprehensive}, but also the ability to integrate incomplete evidence, manage uncertainty, and make clinically coherent decisions. To this end, the research community has developed a diverse set of benchmarks to probe different facets of medical reasoning performance. As summarized in Table~\ref{Table 2}, existing benchmarks broadly fall into two categories: medical-specific benchmarks and general-domain benchmarks with medical subsets.

\begin{table}[htbp]
    \centering
    \caption{Summary of Existing Benchmarks. \protect\underline{Underlined} benchmarks represent general-domain datasets containing specific medical subsets.}
    \label{Table 2}
    \renewcommand{\arraystretch}{1.2} 
    \begin{tabular*}{\textwidth}{@{\extracolsep{\fill}}llc r}
        \toprule
        \textbf{Benchmark} & \textbf{Type} & \textbf{Year} & \textbf{Samples} \\
        \midrule
        \href{https://huggingface.co/datasets/qiaojin/PubMedQA}{PubMedQA} \cite{jin2019pubmedqa} & Multiple Choice QA & 2019 & 1,000 \\
        \href{https://huggingface.co/datasets/bigbio/med\_qa}{MedQA (USMLE)} \cite{jin2021disease} & Multiple Choice QA & 2020 & 1,273 \\
        \href{https://huggingface.co/datasets/cais/mmlu}{\protect\underline{MMLU}} \cite{hendrycks2021measuring} & Multiple Choice QA & 2020 & 15,908 \\
        \href{https://huggingface.co/datasets/openlifescienceai/medmcqa}{MedMCQA (Validation)} \cite{pal2022medmcqa} & Multiple Choice QA & 2022 & 4,183 \\
        \href{https://huggingface.co/datasets/FreedomIntelligence/CMB}{CMBExam} \cite{wang2024cmb} & Multiple Choice QA & 2023 & 11,200 \\
        \href{https://huggingface.co/datasets/Idavidrein/gpqa}{\protect\underline{GPQA}} \cite{rein2024gpqa} & Multiple Choice QA & 2023 & 448 \\
        \href{https://huggingface.co/datasets/TIGER-Lab/MMLU-Pro}{\protect\underline{MMLU-Pro}} \cite{mmlu-pro} & Multiple Choice QA & 2024 & 12,032 \\
        \href{https://huggingface.co/datasets/mkieffer/Medbullets}{MedBullets} \cite{chen2025benchmarking} & Multiple Choice QA & 2024 & 308 \\
        \href{https://huggingface.co/datasets/guan-wang/ReDis-QA}{ReDis-QA} \cite{wang2024assessing} & Multiple Choice QA & 2024 & 1,360 \\
        \href{https://huggingface.co/datasets/cais/hle}{\protect\underline{HLE}} \cite{phan2025humanity} & Multiple Choice QA & 2025 & 2,500 \\
        \href{https://huggingface.co/datasets/jdh-algo/JMED}{JMED} \cite{wang2025citrus} & Multiple Choice QA & 2025 & 1,000 \\
        \href{https://huggingface.co/datasets/TsinghuaC3I/MedXpertQA}{MedXpertQA (Text)} \cite{zuo2025medxpertqa} & Multiple Choice QA & 2025 & 2,450 \\
        \href{https://huggingface.co/datasets/super-dainiu/medagents-benchmark}{MedAgentsBench} \cite{anonymous2025medagentsarena} & Multiple Choice QA & 2025 & 862 \\
        
        \href{https://huggingface.co/datasets/HPAI-BSC/CareQA}{Care-QA} \cite{Care-QA} & Multiple Choice / Open Ended QA & 2024 & 8,390 \\
        \href{https://github.com/ji-youn-kim/ehrnoteqa}{EHRNoteQA} \cite{kweon2024ehrnoteqa} & Multiple Choice / Open Ended QA & 2024 & 962 \\
        
        \href{https://huggingface.co/datasets/truehealth/medicationqa}{MedicationQA} \cite{singhal2023large} & Open Ended QA & 2022 & 690 \\
        \href{https://github.com/openai/simple-evals}{HealthBench} \cite{arora2025healthbench} & Open Ended QA & 2025 & 5,000 \\
        
        \href{https://huggingface.co/datasets/AIM-Harvard/MedBrowseComp}{MedBrowseComp} \cite{chen2025medbrowsecomp} & Open Ended QA & 2025 & 605 \\
        
        \href{https://github.com/MAGIC-AI4Med/MedRBench}{MedR-Bench} \cite{qiu2025quantifying} & Open Ended QA & 2025 & 1453 \\

        \href{https://ehrshot.stanford.edu/}{EHRSHOT} \cite{wornow2023ehrshot} & Clinical Risk Prediction & 2023 & 6,739 \\
        
        \href{https://github.com/paulhager/MIMIC-Clinical-Decision-Making-Dataset}{MIMIC-CDM} \cite{hager2024evaluation} & Clinical Decision Making & 2024 & 2,400 \\
        
        \href{https://github.com/MAGIC-AI4Med/EHR-R1}{EHR-Bench} \cite{liao2025ehr} & Clinical Decision Making / Risk Prediction & 2025 & 21,000 \\
        \href{https://huggingface.co/datasets/ncbi/MedCalc-Bench-v1.0}{MedCalc-Bench} \cite{khandekar2024medcalc} & Numerical Calculation & 2024 & 1,047 \\

        \href{https://arxiv.org/abs/2508.20148}{PHA-Bench} \cite{heydari2025anatomy} & Personal Health Management & 2025 & 7,000+ \tnote{*}  \\
        \bottomrule
    \end{tabular*}

\end{table}

\noindent \textbf{Medical-specific Benchmarks.}
A foundational class of medical reasoning benchmarks is derived from standardized medical licensure and entrance examinations, which serve as proxies for broad professional competency. Representative examples include MedQA, based on the United States Medical Licensing Examination (USMLE) \cite{jin2021disease}, MedMCQA \cite{pal2022medmcqa}, and CMBExam \cite{wang2024cmb}, which collect multiple-choice questions (MCQs) from established exams in the United States, India, and China, respectively. These benchmarks have played a central role in assessing knowledge-intensive medical reasoning, but are primarily limited to static question answering under fully specified contexts.
To address these limitations, recent benchmarks have sought to increase task difficulty, realism, and clinical relevance. MedBullets \cite{chen2025benchmarking} and MedXpertQA \cite{zuo2025medxpertqa} target specialist-level reasoning by drawing on questions resembling advanced board examinations (e.g., USMLE Step~2/3). JMED \cite{wang2025citrus} introduces more realistic clinical case distributions and expanded answer sets to reduce shortcut learning and random guessing, while Care-QA \cite{Care-QA} moves beyond MCQs to open-ended clinical assessment. ReDis-QA \cite{wang2024assessing} further focuses on abductive reasoning in rare disease diagnosis, where evidence is sparse and diagnostic ambiguity is high.

Beyond exam-style assessment, an expanding body of benchmarks evaluates practical clinical reasoning and agentic capabilities. For evidence retrieval and synthesis, PubMedQA \cite{jin2019pubmedqa} assesses reasoning over biomedical literature, while MedBrowseComp \cite{chen2025medbrowsecomp} evaluates an agent’s ability to browse, retrieve, and compare information across heterogeneous web sources. To assess patient-facing reasoning and clinical communication, MedicationQA \cite{singhal2023large} and HealthBench \cite{arora2025healthbench} focus on the safety and quality of responses to open-ended consumer health queries and multi-turn clinical conversations.
Several benchmarks target specific reasoning challenges observed in LLMs. MedCalc-Bench \cite{khandekar2024medcalc} isolates numerical reasoning and medical calculation accuracy, while EHRNoteQA \cite{kweon2024ehrnoteqa} evaluates information extraction and synthesis from electronic health records. Moving beyond extraction, EHRSHOT \cite{wornow2023ehrshot} emphasizes longitudinal clinical risk prediction, and MIMIC-CDM \cite{hager2024evaluation} introduces interactive settings that require active information gathering. EHR-Bench \cite{liao2025ehr} consolidates these challenges into a unified suite of 42 decision-making and risk prediction tasks grounded in real-world clinical data.

Recent benchmarks increasingly conceptualize medical reasoning as a sequential and interactive decision-making process. PHA-Bench \cite{heydari2025anatomy}, built on the WEAR-ME dataset, evaluates agents across ten tasks spanning clinical reasoning, health coaching, and data analysis, supported by over 7,000 human annotations. MedAgentsBench \cite{anonymous2025medagentsarena} explicitly targets multi-agent collaboration in solving complex medical problems. MedR-Bench \cite{qiu2025quantifying} leverages 1,453 structured patient cases from PMC Open Access to evaluate reasoning across examination recommendation, diagnosis, and treatment planning. MedChain \cite{liu2025medchain} introduces a sequential benchmark comprising 12,163 cases that enforces a five-stage clinical workflow, uniquely employing a ``Patient Agent'' to simulate real-time doctor--patient dialogue and requiring models to actively elicit scattered information through interaction rather than reasoning over static inputs.

\noindent \textbf{General-Domain Benchmarks.}
While medical-specific benchmarks directly probe domain knowledge and task competence, general-domain benchmarks with dedicated medical subsets are increasingly employed to evaluate the underlying reasoning capabilities of LLMs \cite{nori2023can, singhal2023large}. Such benchmarks typically operate at a higher level of abstraction and are less susceptible to contamination from medical training corpora, thereby offering a complementary perspective on reasoning generalization and transfer \cite{rein2024gpqa, phan2025humanity}.
Among these, MMLU \cite{hendrycks2021measuring} remains a widely adopted reference point, with its six clinically and biologically relevant subtasks (e.g., Professional Medicine and Anatomy) serving as a coarse-grained indicator of general medical intelligence. MMLU-Pro \cite{mmlu-pro} extends this framework by introducing more challenging questions and an expanded answer space (ten options), thereby increasing the difficulty of reasoning and reducing the effectiveness of superficial heuristics. To explicitly target deep scientific reasoning beyond surface-level retrieval, GPQA \cite{rein2024gpqa} presents a “Google-proof” evaluation, with its biology and chemistry subsets posing expert-level problems that require multi-step inference and conceptual integration.
More recently, HLE \cite{phan2025humanity} has been proposed to address performance saturation on existing benchmarks, offering an expert-level evaluation suite designed to stress-test the upper bounds of model reasoning. A substantial portion of HLE is devoted to medicine and biology, positioning it as a challenging probe of advanced reasoning ability under conditions where factual lookup alone is insufficient.

\subsection{Evaluation Paradigms for Medical Reasoning}

Given the diversity of benchmarks and task formats, a central question is how model performance should be evaluated in a manner that meaningfully reflects medical reasoning ability. Existing work primarily adopts three evaluation paradigms: rule-based evaluation, human-based evaluation, and LLM-based evaluation.

\noindent \textbf{Rule-Based Evaluation.}
Rule-based evaluation relies on automated metrics tailored to specific task formulations. Accuracy remains the standard metric for multiple-choice benchmarks such as MedQA and MMLU, while n-gram overlap measures (e.g., BLEU \cite{papineni2002bleu} and ROUGE \cite{lin2004rouge}) are commonly used for open-ended generation and question answering tasks \cite{kweon2024ehrnoteqa}. The primary advantages of this paradigm are its objectivity, low cost, and high reproducibility, enabling consistent comparison across models and studies \cite{chang2024survey}. However, such metrics predominantly assess final-answer correctness and provide limited visibility into intermediate reasoning processes. Despite this limitation, rule-based evaluation remains the foundational standard for large-scale quantitative benchmarking, offering a reliable but coarse-grained signal of domain competence.

\noindent \textbf{Human-Based Evaluation.}
Human evaluation is widely regarded as the gold standard for assessing clinical correctness, safety, and real-world utility. In this paradigm, expert clinicians review model outputs against multi-dimensional criteria, including factual accuracy, reasoning validity, and potential for harm \cite{awasthi2025human, zheng2025large, esmaeilzadeh2025ethical}. Early large-scale efforts such as Med-PALM and Med-PALM~2 \cite{singhal2023large, singhal2025toward} established rigorous human evaluation protocols, employing panels of physicians to assess long-form medical responses. Subsequent work, including Med-Gemini \cite{saab2024capabilities}, extended human evaluation to long-context reasoning tasks. AMIE \cite{tu2025towards, mcduff2025towards} further advanced this paradigm through randomized, double-blind studies involving simulated patients and specialist clinicians, enabling realistic assessment of diagnostic dialogue. More recently, PHA \cite{heydari2025anatomy} broadened the scope of human evaluation by incorporating multiple stakeholder perspectives, including clinicians, data scientists, and health coaches, validating over 7,000 interactions for medical accuracy, usability, and robustness. Collectively, these studies underscore the irreplaceable role of human evaluation in verifying clinical safety and reasoning quality. Nevertheless, the substantial cost and limited scalability of expert review constitute a persistent bottleneck for rapid model iteration \cite{chang2024survey}.

\noindent \textbf{LLM-Based Evaluation.}
To alleviate the scalability constraints of human evaluation, recent work increasingly adopts large language models as surrogate evaluators. In this \emph{LLM-as-a-Judge} paradigm, a strong reference model (e.g., GPT-5) is used to score or rank the outputs of candidate models, aiming to approximate human judgments at significantly lower cost. Systems such as HuatuoGPT-o1 \cite{chen2412huatuogpt} and Lingshu \cite{xu2025lingshu} demonstrate the feasibility of using LLM-based judges to assess open-ended medical question answering and reasoning quality. To improve reliability, later benchmarks incorporate more structured evaluation protocols. HealthBench \cite{arora2025healthbench} employs clinician-authored rubrics to guide judgment, while PHA \cite{heydari2025anatomy} introduces a human-calibrated evaluation paradigm in which the LLM evaluator is explicitly fine-tuned on expert annotations. MedR-Bench \cite{qiu2025quantifying} further extends this approach by deploying an agentic \textit{Reasoning Evaluator} that decomposes model outputs into explicit reasoning steps and cross-verifies them against external knowledge sources.
Despite their scalability and flexibility, LLM-based evaluation methods remain constrained by the capabilities and biases of the judge model itself. Prior studies have mentioned vulnerabilities such as preference bias, sensitivity to prompt framing, and limited ability to detect subtle medical hallucinations or reasoning errors \cite{zheng2023judging, szymanski2025limitations, krumdick2025no}. As a result, while LLM-based evaluation offers a powerful complementary tool, it cannot yet fully replace human oversight in high-stakes medical reasoning assessment.

\begin{table}[htbp]
    \centering
    \caption{Model performance on existing benchmarks. Background colors denote base LLM families, with each medical LLM inheriting the background color of its corresponding base model. \textbf{Bold} indicates the best overall score; \protect\underline{underlining} marks the best medical LLM within each base model group.}
    \label{tab:model_performance}
    \renewcommand{\arraystretch}{1.2}
    \resizebox{\textwidth}{!}{%
    \begin{tabular}{lcccccccccc}
        \toprule
        \textbf{Model} & \textbf{MedQA} & \textbf{MedMCQA} & \textbf{PubMedQA} & \textbf{GPQA} & \textbf{JMED} & \textbf{ReDis-QA} & \textbf{MedXpertQA} & \textbf{MMLU-Pro} & \textbf{Average} & \textbf{R. I. (\%)} \\
        \midrule
        \multicolumn{11}{c}{\textit{Open-source General LLMs (Base Models)}} \\
        \midrule
        \rowcolor{LightBlue!30} gemma3-4B-it & 0.462 & 0.464 & 0.708 & 0.423 & 0.353 & 0.501 & 0.114 & 0.496 & 0.440 & - \\
        Qwen3-4B-Thinking-2507 & 0.736 & 0.642 & 0.763 & 0.610 & 0.572 & 0.727 & 0.177 & 0.772 & 0.625 & - \\
        \rowcolor{LightYellow} Qwen-2.5-7B-Instruct & 0.582 & 0.562 & 0.748 & 0.474 & 0.580 & 0.667 & 0.131 & 0.623 & 0.546 & - \\
        \rowcolor{Lavender} Meta-Llama-3-8B-Instruct & 0.530 & 0.543 & 0.773 & 0.433 & 0.422 & 0.732 & 0.122 & 0.550 & 0.513 & - \\
        \rowcolor{LightGray!30} Llama-3.1-8B-Instruct & 0.518 & 0.554 & 0.776 & 0.438 & 0.425 & 0.735 & 0.129 & 0.542 & 0.515 & - \\
        Qwen3-8B & \textbf{0.781} & 0.665 & 0.764 & \textbf{0.659} & \textbf{0.668} & 0.778 & 0.182 & \textbf{0.784} & \textbf{0.660} & - \\
        \midrule
        \multicolumn{11}{c}{\textit{Open-source Medical LLMs (vs. Base General LLM)}} \\
        \midrule
        \rowcolor{LightBlue!30} Medgemma 4B & \protect\underline{0.597} & \protect\underline{0.546} & \protect\underline{0.754} & \protect\underline{0.436} & \protect\underline{0.417} & \protect\underline{0.680} & \protect\underline{0.147} & \protect\underline{0.552} & \protect\underline{0.516} & \protect\underline{\textcolor{ForestGreen}{+17.27}} \\
        \rowcolor{LightYellow} ClinicalGPT-R1-Qwen-7B $^{\tnote{\bf a}}$ & 0.431 & 0.526 & 0.670 & 0.462 & 0.078 & 0.649 & 0.112 & 0.559 & 0.436 & \textcolor{red}{-20.15} \\
        \rowcolor{LightYellow} HuatuoGPT-o1-7B & 0.664 & 0.643 & \protect\underline{0.788} & \protect\underline{0.479} & \protect\underline{0.565} & 0.710 & 0.150 & 0.670 & 0.584 & \textcolor{ForestGreen}{+6.92}\\
        \rowcolor{LightYellow} m1-7B-23K & \protect\underline{0.719} & \protect\underline{0.644} & 0.741 & \protect\underline{0.479} & 0.500 & \protect\underline{0.793} & \protect\underline{0.201} & \protect\underline{0.708} & \protect\underline{0.598} & \protect\underline{\textcolor{ForestGreen}{+9.57}} \\
        \rowcolor{Lavender} Med42-v2-8B & 0.562 & 0.582 & \protect\underline{0.782} & 0.456 & 0.424 & \protect\underline{\textbf{0.839}} & 0.136 & 0.541 & 0.540 & \textcolor{ForestGreen}{+5.29} \\
        \rowcolor{Lavender} Meerkat-8B & \protect\underline{0.661} & \protect\underline{0.587} & 0.764 & \protect\underline{0.474} & \protect\underline{0.431} & 0.787 & \protect\underline{0.148} & \protect\underline{0.576} & \protect\underline{0.554} & \protect\underline{\textcolor{ForestGreen}{+7.87}} \\
        \rowcolor{LightGray!30} Adathink-Med-Llama-8B & 0.581 & 0.572 & 0.780 & 0.433 & 0.399 & 0.778 & 0.162 & 0.584 & 0.536 & \textcolor{ForestGreen}{+4.18} \\
        \rowcolor{LightGray!30} AlphaMed-8B & 0.705 & 0.649 & 0.785 & \protect\underline{0.564} & 0.294 & 0.833 & \protect\underline{\textbf{0.230}} & 0.663 & 0.590 & \textcolor{ForestGreen}{+14.72} \\
        \rowcolor{LightGray!30} BioMed-R1-8B & \protect\underline{0.769} & \protect\underline{\textbf{0.667}} & 0.783 & 0.556 & 0.404 & 0.835 & 0.204 & \protect\underline{0.682} & \protect\underline{0.613} & \protect\underline{\textcolor{ForestGreen}{+19.02}} \\
        \rowcolor{LightGray!30} HuatuoGPT-o1-8B & 0.723 & 0.643 & 0.788 & 0.536 & 0.352 & 0.793 & 0.164 & 0.628 & 0.578 & \textcolor{ForestGreen}{+12.39} \\
        \rowcolor{LightGray!30} Llama-3.1-8B-UltraMedical & 0.698 & 0.622 & \protect\underline{\textbf{0.789}} & 0.462 & \protect\underline{0.447} & 0.815 & 0.162 & 0.634 & 0.579 & \textcolor{ForestGreen}{+12.44} \\
        \rowcolor{LightGray!30} MedReason-8B & 0.653 & 0.605 & 0.733 & 0.487 & 0.355 & 0.754 & 0.218 & 0.622 & 0.553 & \textcolor{ForestGreen}{+7.53} \\
        \rowcolor{LightGray!30} MedS$^3$-8B & 0.684 & 0.639 & 0.781 & 0.495 & 0.414 & \protect\underline{\textbf{0.839}} & 0.171 & 0.659 & 0.585 & \textcolor{ForestGreen}{+13.72} \\
        \bottomrule
    \end{tabular}%
    }
\begin{tablenotes}
\item \textit{Abbreviations}: R.I.: Relative Improvement (compared to the corresponding base general LLM).

\end{tablenotes} 
\end{table}

\section{Cross-Benchmark Evaluation of Medical Reasoning Models}

Despite the rapid proliferation of benchmarks for evaluating medical reasoning in large language models, the field still lacks a unified and standardized evaluation protocol. Existing studies often report results on heterogeneous subsets of benchmarks, with variations in prompting, answer parsing, and evaluation settings, making direct comparison across methods difficult. As a result, the empirical impact of many proposed medical reasoning enhancements remains fragmented and challenging to assess in a holistic and reproducible manner.

To address this gap, we conduct a unified cross-benchmark evaluation that systematically compares representative medical reasoning models and enhancement approaches under a consistent experimental protocol. Rather than introducing new tasks at this stage, our objective is to quantify how much performance improvement existing methods yield on widely used benchmarks when evaluated side by side. In this sense, our evaluation consolidates and partially replicates prior results within a substantially more comprehensive and controlled benchmark suite, enabling clearer attribution of performance gains.

\noindent \textbf{Benchmark Selection.}
We select a diverse set of benchmarks spanning multiple dimensions of medical reasoning. Specifically, MedQA (5-option), PubMedQA, and MedMCQA serve as widely adopted standard benchmarks; the medical subsets of MMLU-Pro and GPQA probe higher-level and domain-general reasoning; ReDis-QA evaluates generalization to rare disease diagnosis; MedXpertQA targets expert-level clinical reasoning; and JMED examines performance in a Chinese-language clinical setting. Together, these benchmarks cover a broad spectrum of medical knowledge, reasoning complexity, and linguistic contexts.

\noindent \textbf{Model Selection and Evaluation Protocol.}
To assess the effectiveness of training-based approaches for medical reasoning, we evaluate a collection of open-source medical LLMs alongside their corresponding general-purpose base models across all selected benchmarks. Performance gains are measured as relative improvements over the base models, enabling a controlled assessment of the value contributed by domain-specific adaptation. To ensure fair horizontal comparison and practical reproducibility, we restrict our analysis to open-source models with fewer than 10B parameters. In addition to widely used backbones such as Qwen-2.5-7B-Instruct and Llama-3.1-8B-Instruct, we include the recently released Qwen3-4B-Thinking-2507 and Qwen3-8B \cite{yang2025qwen3technicalreport} to examine how advances in general-purpose LLMs translate to medical reasoning benchmarks. Detailed experimental settings are provided in Appendix~\ref{Appendix 1}.

\noindent \textbf{Experimental Results and Analysis.}
As summarized in Table~\ref{tab:model_performance}, domain-specific adaptation generally leads to measurable improvements in medical reasoning performance. Models such as MedGemma-4B and BioMed-R1-8B achieve relative gains of approximately 17\% and 20\% over their respective base models. Notable exceptions include ClinicalGPT-R1, where performance degradation appears to stem from weight irregularities (see Appendix~\ref{Appendix B}). Performance varies substantially across benchmarks: most models struggle on the Chinese-language benchmark JMED, likely reflecting the English-centric nature of most training corpora, whereas Qwen-based models (e.g., HuatuoGPT-o1 and m1) benefit from stronger multilingual foundations. On the highly challenging MedXpertQA benchmark, AlphaMed-8B attains the best performance, highlighting the potential advantages of reinforcement learning for expert-level medical reasoning.

Beyond domain adaptation effects, our results reveal the rapid progress of general-purpose LLMs. The untuned Qwen3-8B model achieves the highest average performance across benchmarks, and even the smaller Qwen3-4B-Thinking model outperforms several larger, domain-specialized medical LLMs. These findings suggest that improvements in general reasoning capability can rival—and in some cases surpass—gains from medical-specific adaptation. 
This trend underscores that improvements in general reasoning capability can rival, and in some cases surpass, gains from domain-specific adaptation, emphasizing the need to continually reassess medical reasoning pipelines in light of rapidly evolving foundation models.